%% file: main.tex
\documentclass[10pt,twocolumn,letterpaper]{article}
\usepackage{iccv}

\input{packages}

\input{macros}

\iccvfinalcopy %
\def\iccvPaperID{0000} %

\ificcvfinal\fi

\begin{document}

\title{Controllable Visual-Tactile Synthesis}

\author{Ruihan Gao
\qquad \qquad
Wenzhen Yuan
\qquad \qquad
Jun-Yan Zhu\\
\vspace{1mm}
Carnegie Mellon University
}

\twocolumn[{
\renewcommand\twocolumn[1][]{#1}
\maketitle
\input{figText/teaser.tex}
}]

\thispagestyle{empty}
\input{sections/0_abstract}
\input{sections/1_introduction}
\input{sections/2_related_work}
\input{sections/3_dataset}

\input{sections/4_method}
\input{sections/5_exp}

\input{sections/6_conclusion}

\input{sections/7_ack}

{\small
\bibliographystyle{ieee_fullname}
\bibliography{main}
}
\appendix
\clearpage
\input{sections/8_supplement}

\end{document}

%% file: packages.tex
\usepackage[pagebackref=true,breaklinks=true,colorlinks,bookmarks=false]{hyperref}

\usepackage{graphicx}
\usepackage{amsmath}
\usepackage{amssymb}
\usepackage{booktabs} 
\usepackage{times}

\usepackage[capitalize]{cleveref}
\crefname{section}{Sec.}{Secs.}
\Crefname{section}{Section}{Sections}
\Crefname{table}{Table}{Tables}
\crefname{table}{Tab.}{Tabs.}
\Crefname{figure}{Figure}{Figures}

\usepackage[font=small]{caption}
\usepackage{color,xcolor}
\usepackage{epsfig}
\usepackage{comment}
\usepackage{pifont}
\usepackage{microtype}
\usepackage{arydshln}
\usepackage{tabularx}
\usepackage{adjustbox}
\usepackage{array}
\usepackage{colortbl}
\usepackage{float,wrapfig}
\usepackage{hhline}
\usepackage{multirow}
\usepackage[percent]{overpic}
\usepackage{stmaryrd}
\usepackage{breqn}
\usepackage{bm}
\usepackage{nicefrac}
\usepackage{dsfont}
\usepackage{changepage}
\usepackage{extramarks}
\usepackage{fancyhdr}
\usepackage{soul}
\usepackage{xspace}
\usepackage{algorithm}
\usepackage{algorithmicx}
\usepackage[algo2e]{algorithm2e} %
\usepackage{algpseudocode}
\usepackage{enumitem}
\usepackage[title]{appendix}
\usepackage{balance}
\usepackage{booktabs}
\usepackage{siunitx}

\usepackage{subcaption} %

%% file: macros.tex
\definecolor{MyPurple}{RGB}{111,0,255}

\def \mywebsitelink{https://visual-tactile-synthesis.github.io/}

\def \VcGAN{\mathcal{L}_{\text{cGAN (visual)}}}
\def \TcGAN{\mathcal{L}_{\text{cGAN (tactile)}}}
\def \Vrec{\mathcal{L}_{\text{rec (visual)}}}
\def \Trec{\mathcal{L}_{\text{rec (tactile)}}}

\def \datasetName{\texttt{TouchClothing}\xspace}

\def\xinput{x} %
\def\S{x} %
\def\I{y_I} %
\def\T{y_T} %
\def\Sp{x^p} %
\def\Ip{y_I^p} %
\def\Tp{y_T^p} %

\def\GI{G_I} %
\def\GT{G_T} %
\def\DI{D_I} %
\def\DT{D_T} %
\def\DA{D_{\small \text{clip}}} %

\def\gx{g_x} %
\def\gy{g_y} %

\newcommand{\E}{\mathbb{E}\xspace}

\newcommand{\reffig}[1]{Figure~\ref{fig:#1}}
\newcommand{\refsec}[1]{Section~\ref{sec:#1}}
\newcommand{\refapp}[1]{Appendix~\ref{sec:#1}}
\newcommand{\reftbl}[1]{Table~\ref{tbl:#1}}

\newcommand{\refeq}[1]{Eqn.~\ref{eq:#1}}

\newcommand{\lblfig}[1]{\label{fig:#1}}
\newcommand{\lblsec}[1]{\label{sec:#1}}
\newcommand{\lbleq}[1]{\label{eq:#1}}
\newcommand{\lbltbl}[1]{\label{tbl:#1}}

\newcommand{\ignorethis}[1]{}

\newcommand{\myparagraph}[1]{\vspace{3pt} \noindent \textbf{#1} \ }

\def\1{\bm{1}}

\newcolumntype{L}[1]{>{\raggedright\let\newline\\\arraybackslash\hspace{0pt}}m{#1}}
\newcolumntype{C}[1]{>{\centering\let\newline\\\arraybackslash\hspace{0pt}}m{#1}}
\newcolumntype{R}[1]{>{\raggedleft\let\newline\\\arraybackslash\hspace{0pt}}m{#1}}

\newcommand{\ignore}[1]{}

\makeatletter
\DeclareRobustCommand\onedot{\futurelet\@let@token\@onedot}
\def\@onedot{\ifx\@let@token.\else.\null\fi\xspace}

\makeatother

%% file: figText/teaser.tex
\begin{center}
    \centering
    \captionsetup{type=figure}
    \includegraphics[width=\textwidth]{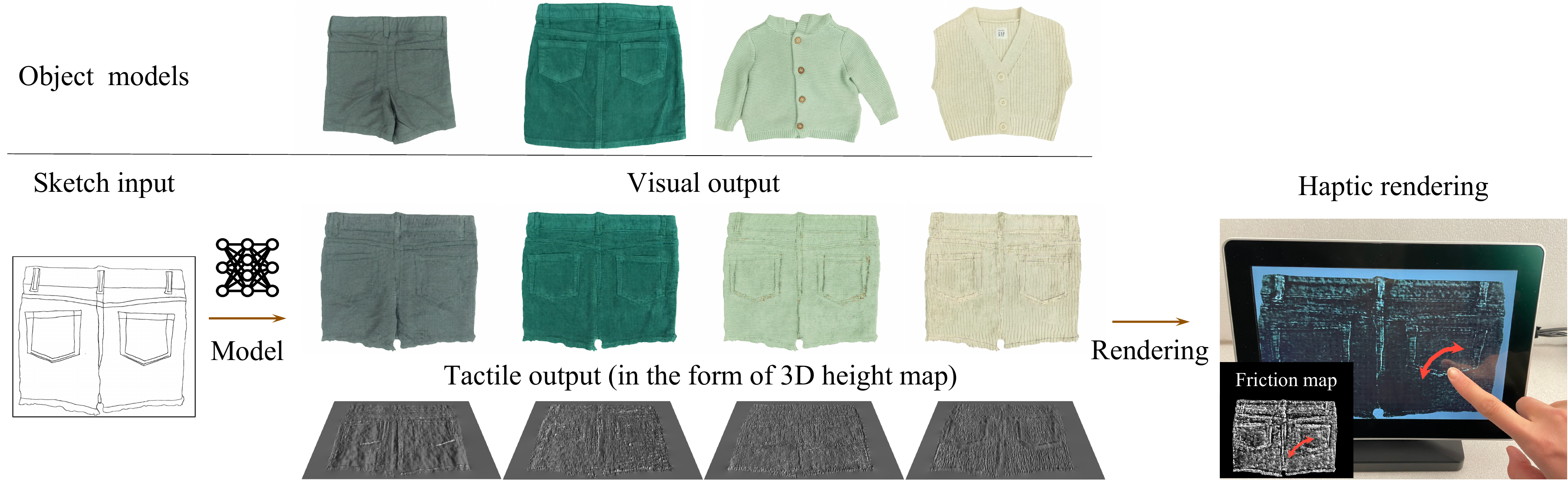} 
    \captionof{figure}{Our method synthesizes visual and tactile outputs from an input sketch and renders the object on a haptic device (e.g., TanvasTouch screen~\cite{tanvas}). Our system allows users to \textit{see} the visual pattern and \textit{feel} the material texture at their fingertips simultaneously. Please see our data capture and user interaction demos at our project page~\url{\mywebsitelink}.
    }
    \lblfig{teaser}
    \vspace{10pt}
\end{center}

%% file: sections/0_abstract.tex
\begin{abstract}
Deep generative models have various content creation applications such as graphic design, e-commerce, and virtual Try-on. However, current works mainly focus on synthesizing realistic visual outputs, often ignoring other sensory modalities, such as touch, which limits physical interaction with users.
In this work, we leverage deep generative models to create a multi-sensory experience where users can \textit{touch} and \textit{see} the synthesized object when sliding their fingers on a haptic surface. The main challenges lie in the significant scale discrepancy between vision and touch sensing and the lack of explicit mapping from touch sensing data to a haptic rendering device. 
To bridge this gap, we collect high-resolution tactile data with a GelSight sensor and create a new visuotactile clothing dataset. We then develop a conditional generative model that synthesizes both visual and tactile outputs from a single sketch. 
We evaluate our method regarding image quality and tactile rendering accuracy. 
Finally, we introduce a pipeline to render high-quality visual and tactile outputs on an electroadhesion-based haptic device for an immersive experience, allowing for challenging materials and editable sketch inputs.
\end{abstract}

%% file: sections/1_introduction.tex
\vspace{-12pt}
\section{Introduction}
\lblsec{intro}

The past few years have witnessed significant progress in content creation powered by deep generative models~\cite{karras2019style,rombach2022high} and neural rendering techniques~\cite{mildenhall2020nerf,Tewari2020NeuralSTAR}. 
Recent works can synthesize realistic images with various user controls, such as user sketches~\cite{isola2017image}, text prompts~\cite{ramesh2022hierarchical}, and semantic maps~\cite{park2019semantic}.  
However, most works focus on synthesizing \emph{visual} outputs, ignoring other sensory outputs such as  touch.

In real life, humans use vision and touch to explore objects. 
When shopping for clothing, we look at them to perceive their shape and appearance and touch them to anticipate the experience of wearing them. 
A single touch can reveal the material's roughness, hardness, and local geometry.  
Multi-modal perceptual inputs enable humans to obtain a more comprehensive understanding of the target objects, enhancing user experiences, such as online shopping and quick prototyping. Moreover, it opens up new possibilities for content creation, such as touchable VR and movies. 

In this work, we aim to expand the capability of content creation. 
We introduce a new problem setting, \emph{controllable visual-tactile synthesis}, for synthesizing high-resolution images and haptic feedback outputs from user inputs of a sketch or text. 
Our goal is to provide a more immersive experience for humans when exploring objects in a virtual environment. 

\looseness=-1
Visual-tactile synthesis is challenging for two reasons. 
First, existing generative models struggle to model visual and tactile outputs jointly due to the dramatic differences in perception scale: vision provides a global sense of our surroundings, while touch offers only a narrow scale of local details. 
Second, there do not exist data-driven end-to-end systems that can effectively render the captured tactile data on a haptic display, as existing haptic rendering systems heavily rely on manually-designed haptic patterns~\cite{birnholtz2015feeling,beheshti2019supporting,klatzky2019detection,sadia2022exploration}.

To address the challenges, we introduce a haptic material modeling system based on surface texture and topography.
We first collect the high-resolution surface geometry of target objects with a high-resolution tactile sensor GelSight~\cite{yuan2017gelsight,wang2021gelsight} as our training data. 
To generate visual-tactile outputs that can render materials based on user inputs, we propose a new conditional adversarial learning method that can learn from multi-modal data at different scales. Different from previous works~\cite{isola2017image,wang2018pix2pixHD}, our model learns from dense supervision from visual images and sparse supervision from a set of sampled local tactile patches. 
During inference, we generate dense visual and tactile outputs from a new sketch design.
We then render our models' visual and tactile output with a TanvasTouch haptic screen~\cite{tanvas}. The TanvasTouch device displays the visual output on a regular visual screen and uses electroadhesion techniques~\cite{shultz2018application} to render the force feedback of different textures according to a friction map. Humans can feel the textures as a changing friction force distribution when sliding their fingers on the screen~\cite{burns2021spatial}.

We collect a spatially aligned visual-tactile dataset named \datasetName that contains 20
pieces of clothing, including pants and shirts, with diverse materials and shapes. 
We evaluate our model regarding image quality and perceptual realism with both automatic metrics and user study. 
Experimental results show that our method can successfully integrate the global structure provided by the sketch and the local fine-grained texture determined by the cloth material, as shown in \reffig{teaser}. 
Furthermore, we demonstrate sketch- and text-based editing applications enabled by our system to generate new clothing designs for humans to \textit{see} and \textit{feel}.
Our code and data are available on our website~\url{\mywebsitelink}.

%% file: sections/2_related_work.tex
\section{Related Work}
\lblsec{related}

\myparagraph{Vision and touch.} 
\looseness=-1
Multimodal perception and learning using vision and touch inputs have been shown effective for several computer vision and robotics applications, such as estimating material proprieties~\cite{yuan2016estimating,yuan2017shape, yuan2015measurement,yuan2017connecting}, object grasping and manipulation~\cite{li2014localization,calandra2017feeling,calandra2018more,wi2022virdo,tian2019manipulation,zhang2021dynamic,luo2021learning}, object recognition~\cite{lin2019learning,sundaram2019learning}, future frame prediction~\cite{yang2022touch}, and representation learning for downstream tasks~\cite{kerr2022learning,lee2018making,yang2022touch}.  
While most existing works focus on improving recognition and learning systems, we aim to synthesize visual-tactile outputs for content creation and VR applications. Several recent works learn to predict tactile outputs given visual inputs~\cite{li2019connecting,cai2021visual,cai2022gan,cao2023vis2hap}. Rather than predicting one modality from the other, we aim to simultaneously synthesize outputs in both modalities from user sketches and text descriptions. 

\myparagraph{Haptic rendering of textures.}
\looseness=-1
Haptic rendering refers to generating physical signals that simulate the feeling of touch and delivering it to humans, typically involving software for modeling and physical hardware for rendering.
Rendering high-resolution material textures remains a challenge, despite extensive studies on the topic
~\cite{bochereau2018perceptual,culbertson2016importance}.
One branch of works~\cite{romano2011creating,culbertson2012refined,culbertson2013generating} used kinesthetic haptic devices to render single-point temporal signals.
Users feel a vibrating force signal when holding a pen-like stylus and sliding on a plane surface.
The lack of spatial resolution during the rendering limited the feeling of reality for haptic rendering. 
Prior works also proposed to render textures on electroadhesion-based devices~\cite{shin2015data,yim2016data,osgouei2018inverse,bhardwaj2019data}, but they are limited to rendering homogeneous textures or coarse object shapes.
In contrast, we propose to use the TanvasTouch device~\cite{tanvas} to render detailed local geometry and material texture of garment objects. This device creates a programmable spatially distributed friction force using electroadhesion, allowing users to feel the texture by sliding their fingers across the touch screen. Using the new device boosts the user's feeling of reality regarding the textures and local geometries. 

\myparagraph{Deep generative models.} 
\looseness=-1
Prior works~\cite{kingma2014adam,goodfellow2014generative,van2016conditional,dinh2016density,ho2020denoising,song2020score,rombach2022high} have enabled various content creation applications such as text-to-image synthesis~\cite{saharia2022photorealistic,ramesh2022hierarchical,ramesh2021zero,yu2022scaling}, virtual Try-on~\cite{han2018viton,lewis2021tryongan,albahar2021pose}, and style transfer~\cite{zhu2017unpaired,saharia2022palette,liu2017unsupervised}. 
Most existing works focus on generating single-modal \emph{visual} output like images, videos~\cite{ho2022imagen}, and 3D data~\cite{park2019deepsdf}. 
Several unconditional GANs synthesize outputs in two domains, such as images and semantic labels~\cite{azadi2019semantic,zhang2021datasetgan,li2021semantic,tritrong2021repurposing}, or RGBD data~\cite{wang2016generative,noguchi2019rgbd}.
While the above works sample multimodal outputs from latent vectors, they are not controllable. In contrast, our method allows us to control multimodal synthesis according to the user inputs. 

\myparagraph{Image-to-image translation.} 
Various methods have adopted conditional generative models~\cite{goodfellow2014generative}to translate an image from one domain to another~\cite{isola2017image,zhu2017unpaired,huang2021multimodal,mirza2014conditional,saharia2022palette,meng2022sdedit,choi2021ilvr}. 
They are widely used in cross-modal prediction tasks such as sketch-to-photo~\cite{isola2017image,sangkloy2017scribbler} and label-to-image~\cite{wang2018pix2pixHD,park2019semantic,zhu2020sean}. 
In contrast, given user input, our model learns to synthesize outputs in two modalities at different spatial scales. Our method also differs from previous works as we learn to synthesize dense tactile outputs from only sparse supervision. 

%% file: sections/3_dataset.tex
\section{Data Acquisition and Hardware}
\lblsec{data}

\looseness=-1
To develop our multimodal synthesis method, we construct a new spatially aligned visual-tactile dataset, \datasetName, which consists of 20 pieces of garments as shown in \reffig{dataset}.
They cover various fabrics commonly seen in the market, such as denim, corduroy, linen, fleece, and wool. 
This dataset could be useful for online shopping and fashion design applications.
For each garment, we obtain a single 1,280 $\times$ 960 visual image capturing the entire object and $\sim$200 tactile patches (32 $\times$ 32 pixels) sparsely sampled from the object surface. 
We track the 3D coordinates of the sensor's contact area and project them on 2D visual images for spatial alignment.
Finally, we extract the contour as the input sketch for each visual image.
Please find our dataset on the website.
Below we detail our collection process. 
\input{figText/dataset}

\myparagraph{Visual-tactile data collection setup.}
\looseness=-1
\reffig{setup} shows our setup to collect aligned visual-tactile data, where each garment object is fixed on a planar stage with tapes.
We capture a top-down view with a Pi RGB Camera mounted on the top aluminum bar and record hundreds of tactile patches by manually pressing a GelSight sensor~\cite{yuan2017gelsight,wang2021gelsight} at different locations of the object in a grid pattern. 
Our setup enables us to capture diverse patches from each object, including the flat sewing pattern with homogeneous texture, local geometry changes such as pocket edges, and randomly distributed features like flower-shaped decoration.

\myparagraph{GelSight tactile sensor.}
\looseness=-1
The GelSight sensor~\cite{yuan2017gelsight,wang2021gelsight} is a vision-based tactile sensor that uses photometric stereo to measure contact geometry at a high spatial resolution of several tens of micrometers. 
In this paper, we use the GelSight R1.5, modified from Wang et al.~\cite{wang2021gelsight}. It has a sensing area of 32mm $\times$ 24mm ($H \times W$) and a pixel resolution of 320$\times$240, equivalent to 100 micrometers per pixel. 
The sensor outputs an RGB image, which can be converted to the surface gradients and used to reconstruct a 3D height map.

\myparagraph{Visual-tactile correspondence.}
To calculate the relative position of the GelSight sensor with respect to the camera, we attach four Aruco markers to the GelSight and run RANSAC~\cite{fischler_bolles_1981} to track its 3D pose. 
This allows us to project the 3D coordinate of the contact area onto the 2D visual image and to determine the bounding box coordinates of each tactile patch.
Example data are shown in \reffig{sample_data}.

\myparagraph{Tactile data pre-processing and contact mask.}
Each tactile output represents a single touch of the GelSight sensor on the garment, where only a small portion of the sensing area is in contact. We observed noticeable artifacts when training the model with raw data. Instead, we mask out the non-contact region and improve the model using only the contact area. 
Specifically, we downsample the tactile output from 320$\times$240 to 104$\times$78 (about 300 micrometers per pixel) to match the image resolution and then create a contact mask for each tactile patch by thresholding the height map. 
We heuristically determine the threshold to be the 75th percentile of the height map values and apply dilation to avoid false negative detections. 
We sample 32$\times$32 patches based on the contact mask as the final tactile data. We capture roughly 200 patches per clothing,  covering $1/6$ of the image area.

\myparagraph{Sketch image.}
We follow the procedure described in pix2pix~\cite{isola2017image} to obtain sketches from visual images. We first extract coarse contours using the DexiNed network~\cite{poma2020dense} and then manually remove small edges to obtain thin contours.

\input{figText/setup.tex}

\myparagraph{TanvasTouch for haptic rendering.}
TanvasTouch~\cite{tanvas} is a haptic screen that renders a distributed friction map for finger contact. It models the air gap between the screen surface and the human finger as a capacitor. When a human finger slides across it, the varying voltage underneath the screen induces a small current in the finger, which is perceived as a changing friction force. The device takes a grayscale friction map as input to modulate the voltage distribution across the screen. The screen displays visual images and renders haptic signals simultaneously, creating a coupled visual-haptic output.

\input{figText/sample_data.tex}

%% file: figText/dataset.tex
\begin{figure}[t]
\centering
\includegraphics[width=.5\textwidth]{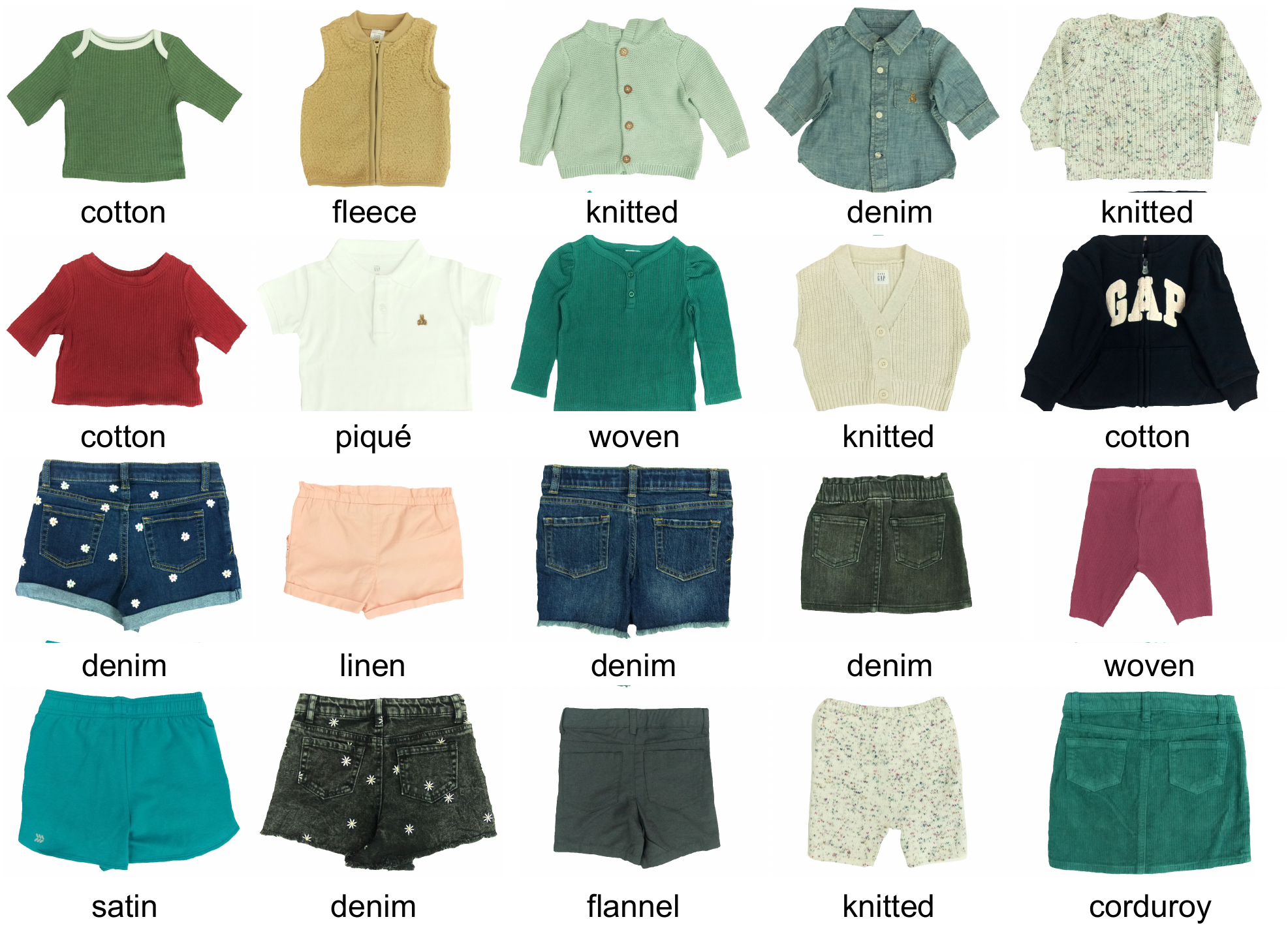}
\caption{\small {\bf Objects in the \datasetName dataset.} Our dataset consists of 20 pieces of clothes with different shapes (shirts, jackets, shorts, pants, etc.) and various fabrics (denim, corduroy, linen, fleece, etc.). Please zoom in to see more details. 
}
\lblfig{dataset}
\end{figure}

%% file: figText/setup.tex
\begin{figure}[t]
\centering
\includegraphics[width=.50\textwidth]{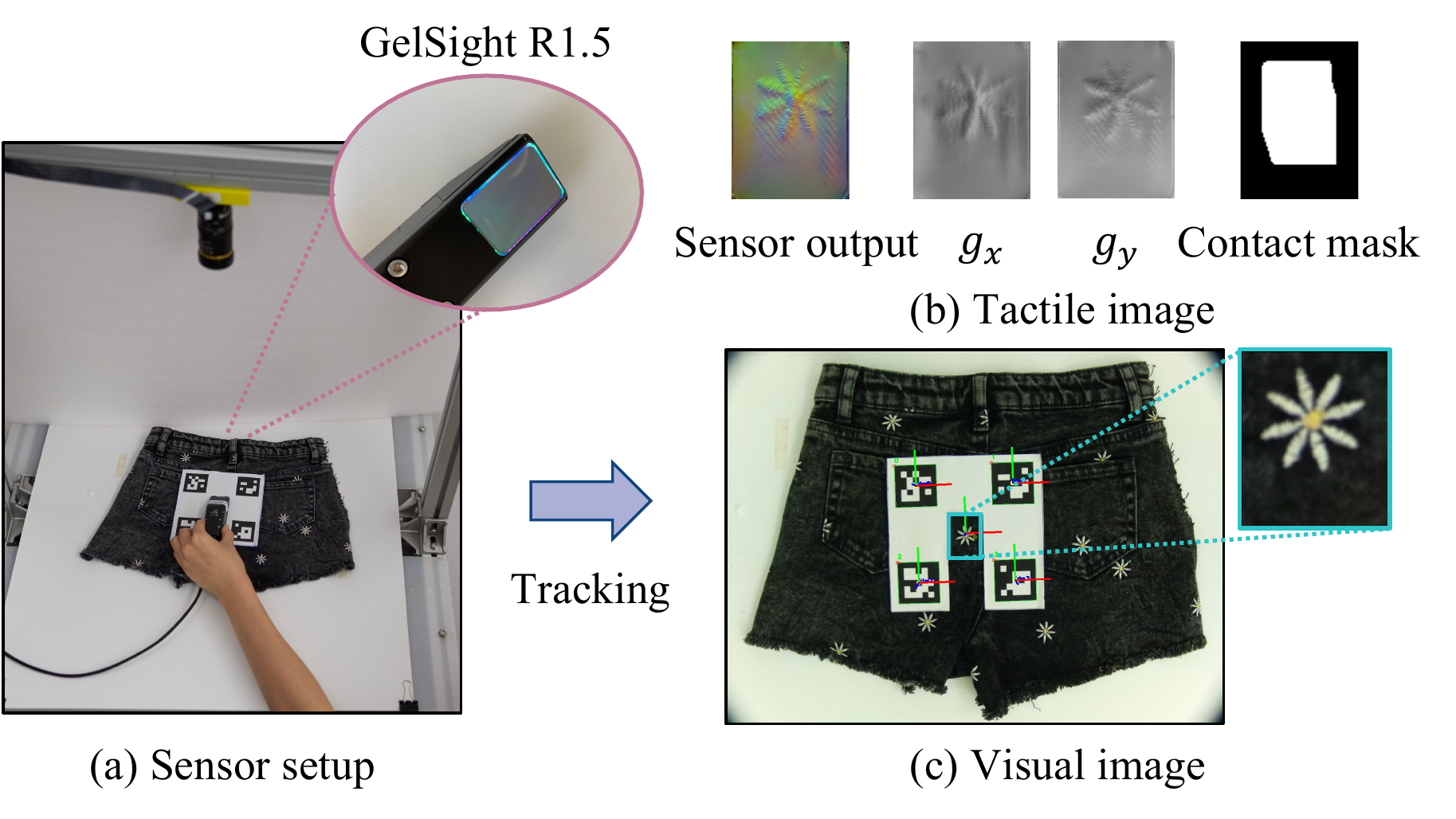}
\caption{\small {\bf Visual-tactile data acquisition setup.} (a) Our setup includes a PiCamera RGB camera, a GelSight R1.5 high-resolution tactile sensor, and Aruco markers to track the relative pose of the sensor. (b) We show the captured tactile data, including the raw sensor output, the derived surface gradients in x and y directions $g_x$ and $g_y$, and the computed contact mask. (c) We locate the bounding box in the visual image corresponding to the tactile data.
}
\lblfig{setup}
\end{figure}

%% file: figText/sample_data.tex
\begin{figure}[t]
\centering
\includegraphics[width=.50\textwidth]{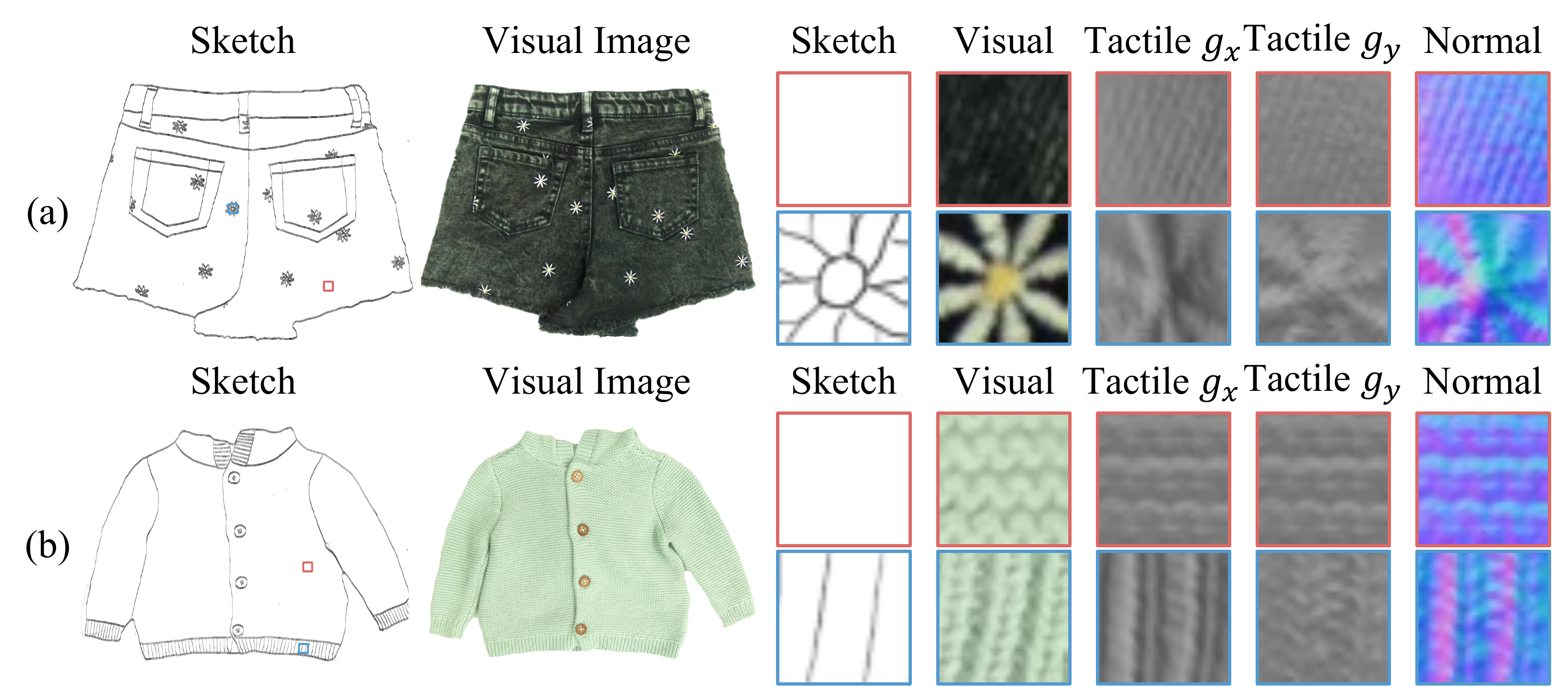}
\caption{\small {\bf Data examples from our \datasetName dataset.} %
For each object, we show the input sketch, the visual image, and two tactile patches. For each tactile patch, we show their corresponding sketch crop, visual crop, and the captured tactile data, including surface gradients in the x and y directions ($g_x$, $g_y$) and surface normal maps. The color-coded bounding boxes in the sketch mark the position of each tactile patch and instantiate the significant scale difference between the visual and tactile data, which makes our conditional synthesis task difficult. 
}
\lblfig{sample_data}
\end{figure}

%% file: sections/4_method.tex
\section{Method}
\lblsec{method}

\input{figText/model.tex}

\looseness=-1
Visual-tactile synthesis is challenging due to the large discrepancy between the receptive field of vision and touch. 
While a camera captures global features of an object, such as color and shape, a touch sensor captures local information within a small patch, such as edges and material texture. 
Existing conditional generative models are not directly applicable as they assume all inputs to be relatively the same scale. 

To address this challenge, we propose a new multi-modal conditional GAN that learns from global visual supervision and sparse local tactile supervision. 
As shown in \reffig{model}, our model synthesizes spatially aligned visual-tactile output given a single sketch. 
We formulate the task in \refsec{formulation} and introduce our learning objective in \refsec{objectives}. 
We describe the network design in \refsec{architecture} and discuss how to render the visual and tactile outputs on the TanvasTouch haptic device in \refsec{haptic}.

\subsection{Visual-Tactile synthesis}
\lblsec{formulation}
We train one object for each object and formulate the visual-tactile synthesis task as a conditional form of single-image generative modeling~\cite{park2020contrastive,shocher2019ingan,shaham2019singan}, which has demonstrated flexible editing ability even though the model is trained on a single image. Specifically, given a single sketch $\S$ of size $H \times W$, where $H$ and $W$ are the image height and width, we aim to learn a function that maps the input sketch $\S$ to two spatially aligned outputs, an RGB visual image $\I$ and a tactile output $\T$.

The sketch $\S$ is a contour map that outlines the object and captures its coarse-scale edge and patterns. For example, in \reffig{sample_data} (a), 
the sketch of a pair of shorts illustrates the overall shape of the shorts, the location of pockets and waistbands, and local decorative patterns. 
In practice, we follow Isola et al.~\cite{isola2017image} to extract a sketch using DexiNed~\cite{poma2020dense} and edge thinning.
\reffig{sample_data} shows examples of the sketch, visual, and tactile images for a pair of shorts and a sweater.

The visual image $\I$ is an RGB image captured by the camera. %
The tactile output $\T=(\gx, \gy)$ is a 2-channel image representing the gradients of the surface in $x$ and $y$ direction. 
They can be converted into surface normal $\mathbf{n}$ using \refeq{compute_normal} and then converted into a height map by Poisson integration~\cite{yuan2017gelsight}.
Since the tactile output is obtained from a calibration network mapping GelSight raw output (RGBXY) to surface gradient $(\gx, \gy)$~\cite{wang2021gelsight}, it is more robust to local noise and position shift in sensor coordinates. It is also less sensitive to integration errors that the height map may suffer after Poisson integration. Therefore, our conditional GAN uses $(\gx, \gy)$ as the tactile output format.

\begin{equation}
\begin{aligned}
\mathbf{n} &= \frac{(\gx, \gy, -1)}{\sqrt{\gx^2 + \gy^2 + 1}}, \quad \gx = \frac{n_x}{n_z}, \quad \gy = \frac{n_y}{n_z}.%
\lbleq{compute_normal}
\end{aligned}
\end{equation}

The generated visual and tactile outputs can be used for applications such as fashion design and haptic rendering. In this work, we render a garment on the TanvasTouch screen, allowing people to simultaneously \emph{see} and \emph{feel} it.

\subsection{Learning Objective}
\lblsec{objectives}

We have two main challenges in this learning task. First, we must learn from dense vision images and sparse tactile supervision while accounting for scale differences. Second, we have limited training data, as we need to learn a synthesis network on a single high-resolution example. To address these challenges, we introduce the following learning objective.

\myparagraph{Visual synthesis loss.}
To synthesize a realistic visual image $\I$ conditional on a user sketch $\xinput$, we optimize the visual generator $\GI$ and visual discriminator $\DI$ to match the conditional distribution of real sketch-image pairs. We optimize the following minimax objective~\cite{isola2017image,mirza2014conditional}:
\begin{equation}
\begin{aligned}
    V(\GI, \DI, \xinput, \I)  &= \mathbb{E}_{\xinput, \I} [\log \DI(\xinput, \I)] \\
    & + \mathbb{E}_{\xinput} [ \log(1 - \DI(\xinput, \GI(\xinput))) ].
\end{aligned}
\lbleq{gans}
\end{equation}
Unfortunately, the above adversarial loss introduces training instability due to our single-image training setting. 
To accommodate the limited dataset size, we use a vision-aided discriminator $\DA$~\cite{kumari2022ensembling} that consists of a frozen CLIP feature extractor~\cite{radford2021learning} and a small trainable MLP head. 
The vision-aided loss can reduce overfitting issues for small-scale datasets and synthesize visual images that better match human perception. 
Our adversarial loss includes: 
\begin{equation}
    \begin{aligned}
        \mathcal{L}_{\text{cGAN}} &= V( \GI, \DI, \xinput, \I) + V( \GI, \DA, \xinput, \I).
    \end{aligned}
\end{equation}
To further stabilize GANs training, we incorporate a reconstruction-based loss. 
Here we use a combination of pixel-wise L1 distance and CNN feature-based perceptual loss (LPIPS)~\cite{zhang2018unreasonable}, as they encourage sharper images~\cite{isola2017image} and higher perceptual similarity to the ground truth.
\begin{equation}
\begin{aligned}
    \mathcal{L}_{\text{rec}}(\GI, \xinput, \I) &= \E_{\xinput,\I} [\mathcal{L}_{\text{LPIPS}}(\I, \GI(\xinput))] \\
    &+ \lambda_{\text{1}} \E_{\xinput,\I}  [\|\I - \GI(\xinput)\|_1], 
    \lbleq{LrecI}
\end{aligned}
\end{equation}
where $\lambda_{\text{1}}$ balances the perceptual loss and L1 loss. 
The final objective function for visual output can be written as follows:  
\begin{equation}
\begin{aligned}
   \mathcal{L}_I &= \mathcal{L}_{\text{cGAN}} + \mathcal{L}_{\text{rec}}. 
\lbleq{objective_I}
\end{aligned}
\end{equation}

\myparagraph{Tactile synthesis loss.}
Unfortunately, we cannot simply use the above loss function to synthesize tactile output, as we no longer have access to the full-size tactile ground truth data. 
Additionally, the vision-aided loss does not apply to tactile data and small patches, as the vision-aided discriminator $\DA$ is pretrained on large-scale natural image collections. 

\looseness=-1
Instead, we learn a full-size tactile generator $\GT$ with supervision from hundreds of tactile patches. Here we denote corresponding (sketch, image, tactile) patches as $(\Sp, \Ip, \Tp)$ at sampled location $p$. 
While the generator $\GT$ synthesizes the full-size tactile output at once, our patch-level discriminator $\DT$ learns to classify whether each patch pair is real or fake, with the following objective: 
\begin{equation}
\begin{aligned}
    & V(\GT, \DT, \xinput, \I, \T)  = \mathbb{E}_{\xinput, \I, \T, p} [\log \DT(\Sp, \Ip, \Tp)] \\
    & + \mathbb{E}_{\xinput, p} [ \log(1 - \DT(\Sp, \GI^p(\xinput), \GT^p(\xinput))) ], 
\end{aligned}
\lbleq{tactilegans}
\end{equation}
where $\GI^p(\xinput)$ and $\GT^p(\xinput)$ denote cropped patches of synthesized visual and tactile outputs. To reduce training memory and complexity, we do not backpropagate the gradients to $\GI$. 

Besides the standard non-saturating GAN objective, we use the feature matching objective~\cite{wang2018pix2pixHD} based on the discriminator's features as the discriminator adapts to the tactile domain better, compared to a pre-trained CLIP model.  In addition, we also add a patch-level reconstruction loss. 
Our final loss for the tactile synthesis branch can be written as follows: 
\begin{equation}
\begin{aligned}
   \mathcal{L}_T = \lambda_{\text{GAN}} V(\GT, \DT, \xinput, \I, \T) + \lambda_{\text{rec}} \mathcal{L}_{\text{rec}}(\GT, \Sp, \Tp). 
\lbleq{objective_T}
\end{aligned}
\end{equation}

\myparagraph{Patch sampling.} We sample two types of patches.  We sample patches with paired ground truth tactile data, for which we can use both reconstruction loss and adversarial loss. 
However, we only have 200 patches for training. To further increase training patches, we also randomly sample patches without paired ground truth. We only try to minimize the second term $\log(1 - \DT(\xinput^p, \GI^p(\xinput), \GT^p(\xinput)))$ of the tactile adversarial loss (\refeq{tactilegans}) as it is only dependent on synthesized patches.

\myparagraph{Full objective.} Our final objective function is 
\begin{equation}
\begin{aligned}
   \GI^*, \GT^* &=\arg \min_{\GI, \GT} \max_{\DI, \DT, \DA}{(\mathcal{L}_I + \mathcal{L}_T)}.
\lbleq{overall_objective}
\end{aligned}
\end{equation}
The weights are chosen using a grid search so that the losses have a comparable scale, and the final values are $\lambda_{\text{1}}=100$, $\lambda_{\text{GAN}}=5$, $\lambda_{\text{rec}}=10$. 
The grid search is done only once for a randomly selected object, and the same parameters are used for all objects in the dataset. 

In \refsec{expr}, we carefully evaluate the role of the adversarial loss and image reconstruction loss regarding the performance of our final model. 

\subsection{Training details}
\lblsec{architecture}

Below we describe our generator and discriminator's network architectures and other training details.

\myparagraph{Network architectures.}
We use a U-Net~\cite{ronneberger2015u} as the backbone of our generator, which splits into two branches, $\GI$ and $\GT$, from an intermediate layer of the decoder. 
This way, the visual and tactile outputs share the same encoding for global structure while maintaining modality-specific details at each pixel location. For discriminators, we use multi-scale PatchGAN~\cite{isola2017image,wang2018pix2pixHD} for both visual discriminator $\DI$ and tactile discriminator $\DT$, since multi-scale PatchGAN has been shown to improve the fine details of results.

\myparagraph{Positional encoding and object masks.}
Since sketches often contain large homogeneous texture areas, we use Sinusoidal Positional Encoding (SPE)~\cite{xu2021positional} to encode the pixel coordinates and concatenate the positional encoding and the sketch at the network input. 
We also extract the object mask and use it to remove the background from the input and output. 
Thus the final input to the network is a masked version of the concatenated sketch and positional encoding features. 
Please refer to our \refapp{more_exp} for more training details. 

\subsection{Haptic rendering}
\lblsec{haptic}
After synthesizing the visual and tactile output, we render them on the TanvasTouch haptic screen using the following rendering pipeline so that users can \emph{see} and \emph{feel} the object simultaneously. 
Specifically, we display the visual image directly on the screen and convert the two-channel tactile output $(g_x, g_y)$ into a grayscale friction map required by TanvasTouch.
As shown by Manuel et al.~\cite{manuel2015coincidence} and Fiesen et al.~\cite{friesen2021building}, humans are sensitive to contours and high-frequency intensity change for surface haptic interpretation. 
Inspired by this, we first compute the squared magnitude of the gradient $z = g_x^2 + g_y^2$, $z\in [0,1]$, then apply non-linear mapping function $z' = \log_{10}(9\times z+1)$, $z'\in [0,1]$ for contrast enhancement, and finally resize it to the TanvasTouch screen size as the final friction map. 
We empirically find this helpful to enhance textures' feeling with electroadhesive force.

%% file: figText/model.tex
\begin{figure*}[t]
\centering
\includegraphics[width=\textwidth]{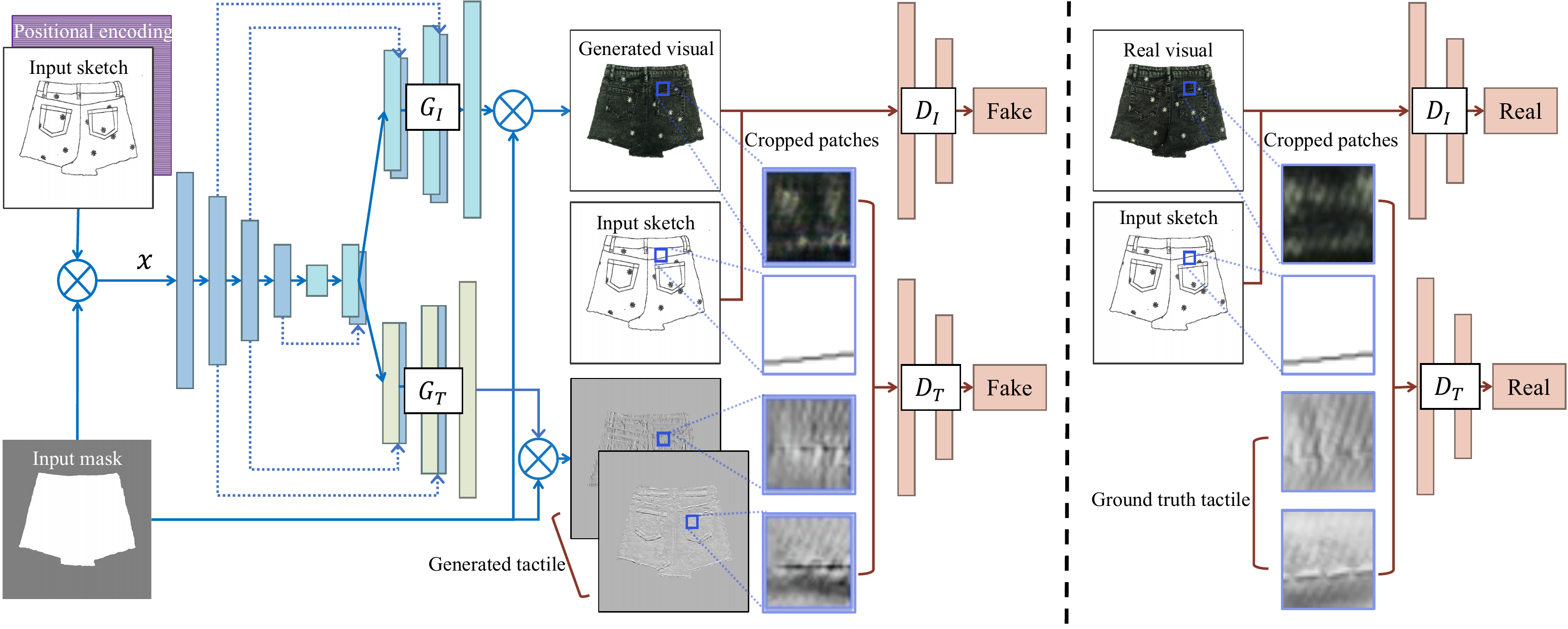}
\caption{\small {\bf Overview.} \emph{Generators}: Given a user sketch, its foreground mask, and positional encoding of the pixel coordinates, we feed them into a two-branch generator.
The two branches share the encoder and the first four layers of the decoders and then split to synthesize visual and tactile results, respectively. 
\emph{Discriminators}: We feed the entire visual image to our visual discriminator $D_I$ and patches to our tactile discriminator $D_T$. $D_I$ is conditional on the sketch, and $D_T$ is conditional on both sketch crops and visual crops.
}

\lblfig{model}
\end{figure*}

%% file: sections/5_exp.tex
\section{Experiment}
\lblsec{expr}

Below we present our main results. Please check out our website for data capture and user interaction videos. 

\looseness=-1
\myparagraph{Evaluation metrics.}
We evaluate our method on the similarity between the synthesized output and the real data of the \datasetName dataset. 
For both visual and tactile output, we report the LPIPS metric~\cite{zhang2018unreasonable} for perceptual realism as prior works~\cite{zhang2018unreasonable,johnson2016perceptual} have shown that the LPIPS metric better matches human perception, compared to PSNR and SSIM~\cite{wang2004image}. 
We also use Single Image Fréchet Inception Distance (SIFID)~\cite{shaham2019singan} for texture similarity, as extensively used in prior works~\cite{shaham2019singan,park2020swapping}. 
Since the dataset only contains one visual image per object, we evaluate LPIPS on seen sketches for visual reconstruction and SIFID on unseen sketches for texture consistency in generalization. 
In addition to automatic metrics, we perform a human preference study.

\myparagraph{Baselines.}
To our knowledge, this paper is the first to study visual-tactile synthesis conditioned on a sketch input. 
Thus we consider image-to-image translation as a similar task and compare our method with several conditional GANs, including pix2pix~\cite{isola2017image}, pix2pixHD~\cite{wang2018pix2pixHD} and GauGAN~\cite{park2019semantic}. 
Pix2pix is one of the most commonly used image translation networks, pix2pixHD uses a coarse-to-fine generator and a multi-scale discriminator to handle high-resolution image synthesis, and GauGAN adopts spatially-adaptive denormalization layers. 
Both pix2pixHD and GauGAN are trained using a perceptual loss, a conditional GAN loss, and a GAN-based feature matching loss.

For baselines, we add two channels for tactile output $\gx$ and $\gy$, increasing the number of output channels from 3 to 5. 
The visual and tactile outputs are fed into two discriminators, both conditioned on the sketch input. 
Since only patch data are available as tactile ground truth, we crop the corresponding region of the sketch and visual images into patches and train the network using sketch-visual-tactile patch pairs. 
We perform the same amount of augmentation as our method. We follow the default parameters in the original works. 
During inference, we feed in the entire sketch image to obtain the full-scale visual and tactile outputs, as the fully convolutional network generalizes to inputs of different sizes.

\myparagraph{Quantitative comparisons.}
As shown in \reftbl{baseline}, our method outperforms all baselines by a large margin in all metrics.
Our method reduces visual LPIPS by more than 50\% and tactile LPIPS by about 30\%. 
Our results depict more realistic and faithful textures, as demonstrated by $5\times$ and $2\times$ lower SIFID for visual and tactile output, respectively. 
This shows the advantage of our method for both visual and tactile synthesis. 
We notice that pix2pix works better than pix2pixHD and GauGAN regarding most metrics. 
This may be because all baselines require paired datasets, and in our case, paired data are low-resolution (32$\times$32), which does not fit the application of pix2pixHD and GauGAN.

\input{figText/baseline.tex}

\myparagraph{Qualitative results.}
\reffig{baseline} provides an example of qualitative comparisons with baselines. 
For each method, the first row shows the full-scale visual output; the second row shows the reconstructed 3D height map; the third row shows some sampled patches in visual, grayscale $\gx$, $\gy$, and derived surface normal formats.
Our method can successfully capture the prominent geometric features, such as pockets and flower-pattern decorations, and the local geometry details of the material textures. 
In contrast, baselines can only capture some prominent geometric features but miss local texture details and generate color artifacts.

\input{tblText/baseline.tex}

\input{tblText/ablation_loss.tex}

\myparagraph{Generation using unseen sketch images.}
Our visual-tactile synthesis model trained on a single sketch image can be generalized to new sketch inputs, allowing users to edit and customize their sketches for fast design and prototyping. 
Since we train one model per object, we show the testing results using sketches of unseen objects in \reffig{swap}.
Each row corresponds to one testing sketch, and each column represents a model trained on one object. 
We visualize results by showing the visual image on the left and the normal map on the right. 
The visual and tactile outputs are well aligned and maintain fine-scale material texture details for each model. 
Our method can adapt to the global geometry information, including the edges and pockets of new sketch inputs.

\input{figText/swap.tex}

\myparagraph{Text-contioned visual-tactile synthesis.}
We also extend our method to synthesizing visual-tactile outputs given both sketches and text prompts. 
We use DALL$\cdot$E2~\cite{ramesh2022hierarchical} to create variations of an original sketch and then feed the edited sketches to our conditional generative models. 
\reffig{dalle} shows examples of text-based synthesis with text prompts.
Even when trained on a single sketch, our model can reasonably generalize to unseen sketches with varying strokes and shapes while capturing the visual and tactile features of the original material. %

\input{figText/dalle.tex}

\myparagraph{Ablation studies.}
We run ablation studies on each loss component to inspect their effects on the training objective. 
\reftbl{ablation_loss} shows that removing either adversarial loss or reconstruction loss for both visual and tactile synthesis together increases LPIPS errors and SIFID metric. 
Qualitatively, we observe overly smooth images after removing adversarial loss and checkerboard artifacts after removing reconstruction loss.
Please see our \refapp{more_results} for more visual results.

\myparagraph{Human Perceptual Study for Visual Images.}
We perform a human perceptual study using Amazon Mechanical Turk (AMTurk). We do a paired test with the question - ``Which image do you think is more realistic?''. 
Each user has five practice rounds followed by 30 test rounds to evaluate our method against pix2pix, pix2pixHD, GauGAN, Ours w/o $\mathcal{L}_{\text{cGAN}}$, and Ours w/o $\mathcal{L}_{\text{rec}}$. 
All samples are randomly selected and permuted, and we collect 1,500 responses. 
As shown in \reffig{user_study_results_visual}, our method is preferred over all baselines, even compared to Ours w/o $\mathcal{L}_{\text{cGAN}}$ and Ours w/o $\mathcal{L}_{\text{rec}}$, which shows the importance of each term. 

\input{figText/user_study_results}
\looseness=-1
\myparagraph{Human Perceptual Study for Haptic Rendering.}
We also perform a human perceptual study to evaluate the perceived fidelity of the generated haptic output, following conventions in prior works~\cite{grigorii2021data,casticco2022usability}. 
We render two different haptic outputs on the TanvasTouch screen side by side with the same ground-truth visuals and ask participants ``Which side do you feel better matches the real object material?''. 
Twenty people, 13 males and 7 females with an average of 24.1 years (SD: 2.1), participated in the experiments. \reffig{user_study_setup} shows an example setup, and more details can be found in our \refapp{more_exp}.
As shown in \reffig{user_study_results_haptic}, participants strongly favor our method over all other baselines (chance is $50\%$). 
$76.7\%$ of the participants prefer our method to pix2pixHD; compared with pix2pix and GauGAN, our method has a larger advantage, winning $79.6\%$ and $84.2\%$ of the participants, respectively. 
It is harder for users to distinguish the ablated models, but our method still beats Ours w/o $\mathcal{L}_{\text{cGAN}}$ and Ours w/o $\mathcal{L}_{\text{rec}}$, by  $52.1\%$ and  $64.3\%$ respectively. 
The user study results are consistent with the quantitative evaluation using various metrics shown in \reftbl{baseline}.

\input{figText/user_study_setup.tex}

%% file: figText/baseline.tex
\begin{figure*}[ht]
\centering
\includegraphics[width=\textwidth]{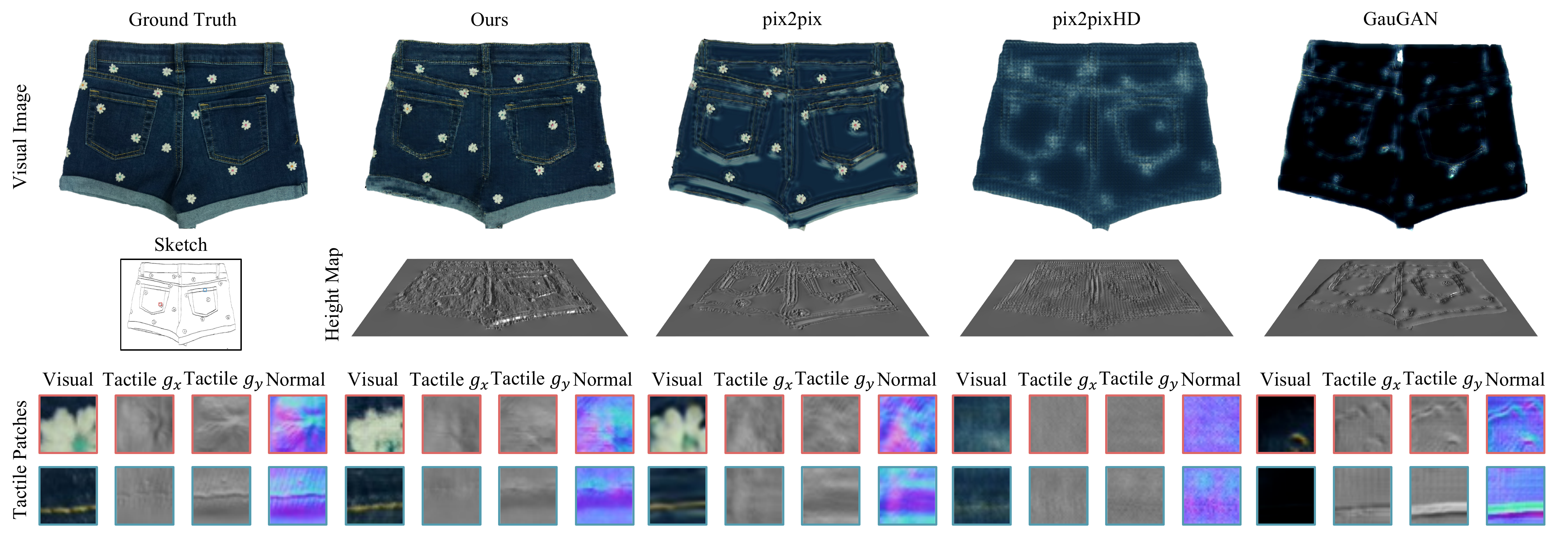}
\caption{\small {\bf Qualitative comparisons with baselines.} We compare our method with the pix2pix~\cite{isola2017image}, pix2pixHD~\cite{wang2018pix2pixHD}, and GauGAN~\cite{park2019semantic}.  For each method, we show the visual output (top) and the rendering of their height maps (middle). 
We also present two zoom-in patches in color-coded bounding boxes (bottom), paired with the visual crop, tactile surface gradients, and the normal map.}

\lblfig{baseline}
\end{figure*}

%% file: tblText/baseline.tex
\begin{table}[t]
\centering
\scalebox{0.8}{
\begin{tabular}{lcccc}
\toprule
\multirow{2}{*}{\textbf{Method} }          & \multicolumn{2}{c}{Visual} & \multicolumn{2}{c}{Tactile} \\
\cmidrule(lr){2-3}\cmidrule(lr){4-5}
 &  \text{LPIPS}$\downarrow$  &  \text{SIFID}$\downarrow$ &  \text{LPIPS}$\downarrow$  & \text{SIFID}$\downarrow$ \\ \midrule
Ours                            & \textbf{0.070}  & \textbf{0.029} & \textbf{\textbf{0.676}} & \textbf{0.104}\\ %
Pix2pix~\cite{isola2017image}   & 0.173  & 0.115 & 1.028  & 0.247 \\ %
Pix2pixHD~\cite{wang2018pix2pixHD}   &  0.161 & 0.289 & 0.753 & 0.458 \\ %
GauGAN~\cite{park2019semantic}  & 0.189 & 0.252 & 1.034 & 0.286  \\ %
\bottomrule
\end{tabular}%
}
\caption{\textbf{Baseline comparisons.} Our method outperforms all baselines regarding both perceptual realism measured by LPIPS~\cite{zhang2018unreasonable} and texture consistency measured by SIFID~\cite{shaham2019singan}.}
\lbltbl{baseline}
\end{table}

%% file: tblText/ablation_loss.tex
\begin{table}[t]
\centering
\scalebox{0.8}{
\begin{tabular}{lccccc}
\toprule
\multirow{2}{*}{\textbf{Method} }          & \multicolumn{2}{c}{Visual} & \multicolumn{2}{c}{Tactile} \\
\cmidrule(lr){2-3}\cmidrule(lr){4-5}
 &  \text{LPIPS}$\downarrow$  &  \text{SIFID}$\downarrow$ &  \text{LPIPS}$\downarrow$  & \text{SIFID}$\downarrow$ \\ \midrule
Ours                            & \textbf{0.070}  & \textbf{0.029} & \textbf{\textbf{0.676}} & \textbf{0.104} \\ %
Ours w/o $\mathcal{L}_{\text{cGAN}}$   & 0.113 & 0.115 & 0.687  & 0.107 \\ %
Ours w/o $\mathcal{L}_{\text{rec}}$     & 0.084& 0.079 & 1.035 & 0.260 \\ %
\bottomrule
\end{tabular}%
}
\caption{\textbf{Ablation study of loss components}.  We compare our full method with two variants: Ours w/o $\mathcal{L}_{\text{cGAN}}$ (w/o conditional GAN losses) and $\mathcal{L}_{\text{rec}}$ (w/o image reconstruction loss). Our full method outperforms these variants regarding multiple metrics. 
}
\lbltbl{ablation_loss}
\end{table}

%% file: figText/swap.tex
\begin{figure}[t]
\centering
\includegraphics[width=0.45\textwidth]{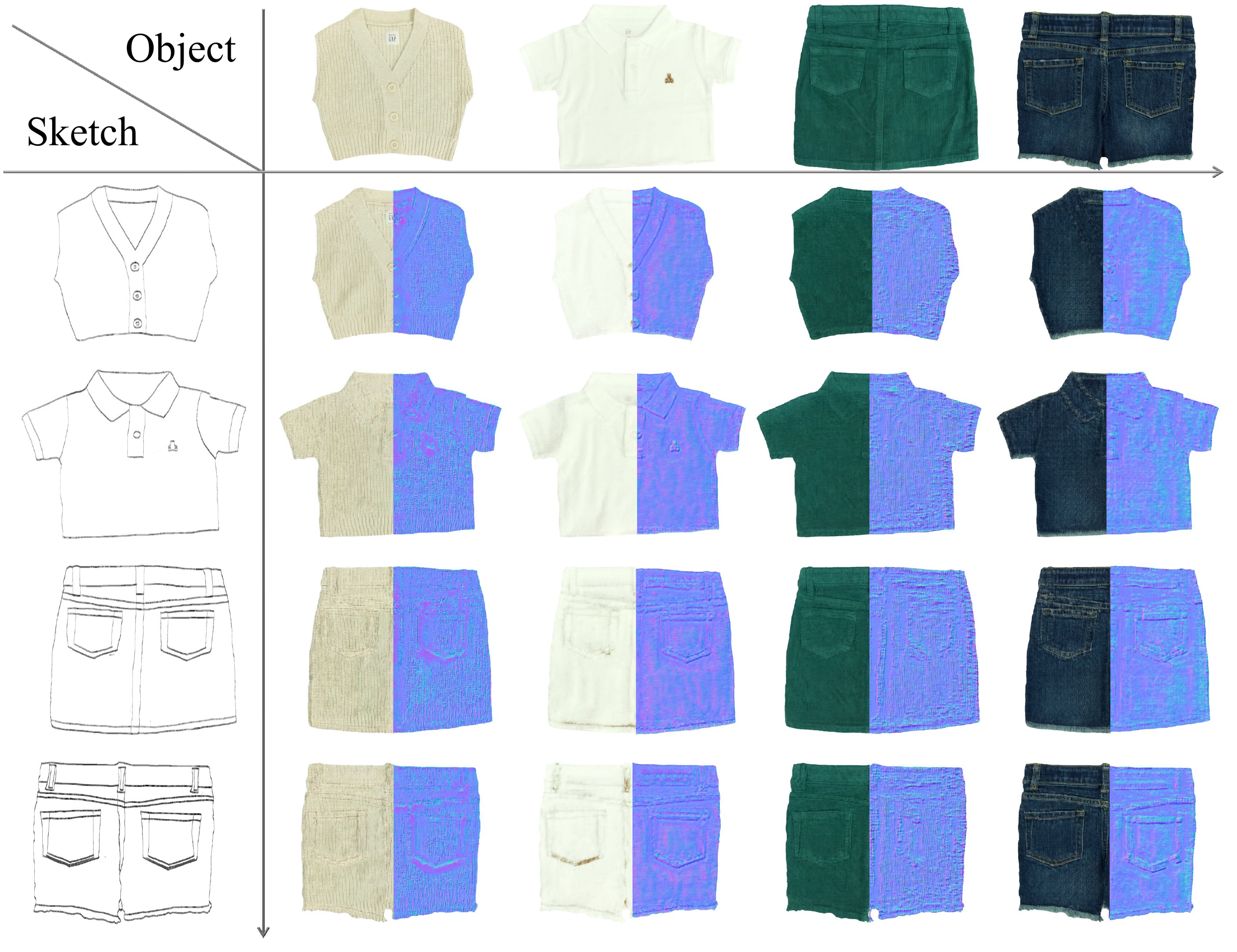}
\vspace{-5pt}
    \caption{\small {\bf Sketch and material swapping.} Our model can synthesize vision and touch images for both known and unseen sketches. For each output, we show visual output at the left half and a normal map of tactile output at the right half.
}
\lblfig{swap}
\vspace{-15pt}
\end{figure}

%% file: figText/dalle.tex
\begin{figure}[t]
\centering
\includegraphics[width=0.5\textwidth]{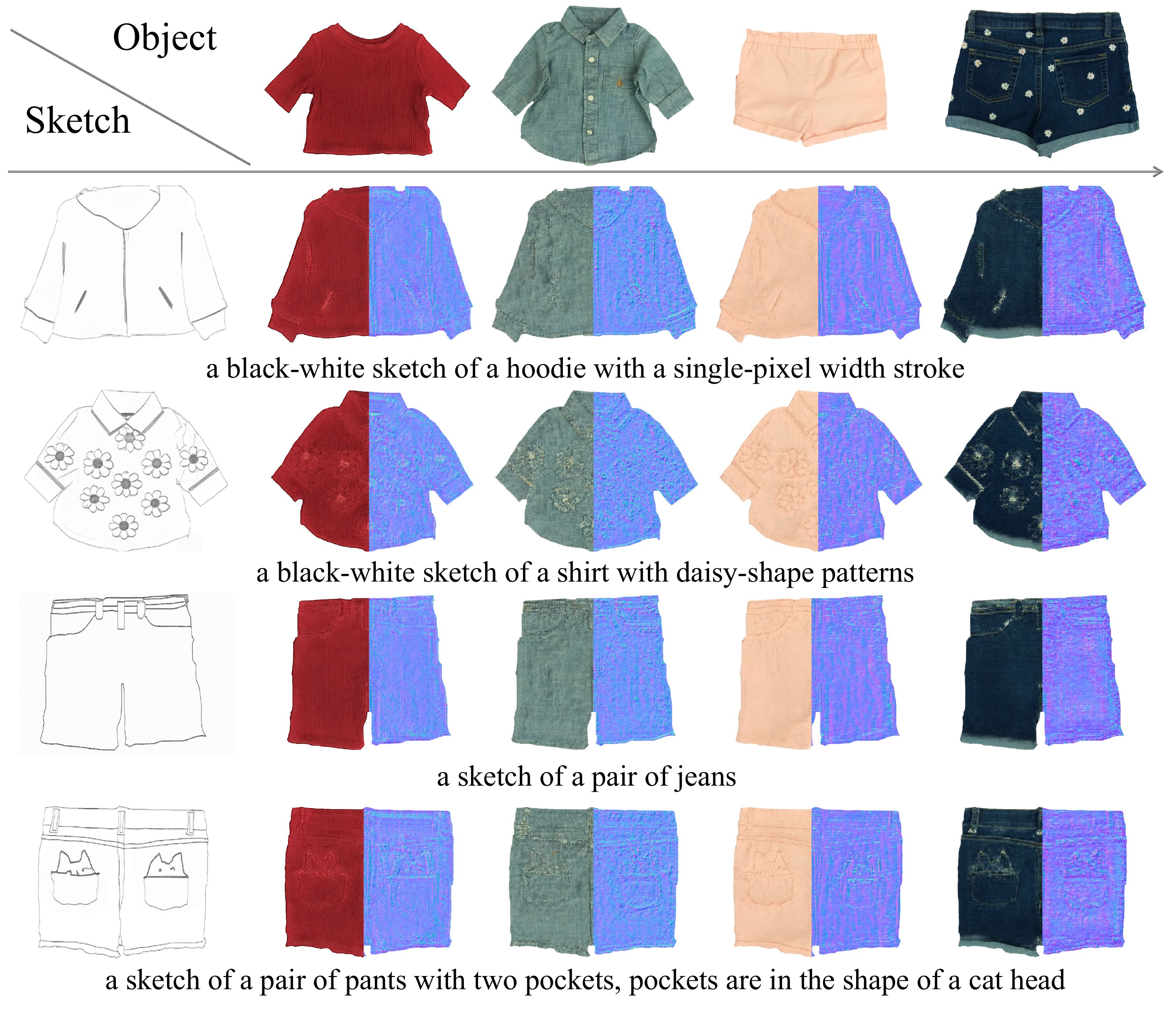}
\caption{{\bf Text-based visual-tactile synthesis.} We use DALL$\cdot$E2~\cite{ramesh2022hierarchical}, a text-to-image model, to modify the original sketch designs via text-based image inpainting. 
We then synthesize both visual (left) and tactile (right) outputs using edited sketches. 
The text prompts are included below each image.
}

\lblfig{dalle}
\end{figure}

%% file: figText/user_study_results.tex
\begin{figure}[t]
  \centering
  \begin{subfigure}[b]{0.48\linewidth}
    \includegraphics[width=\linewidth]{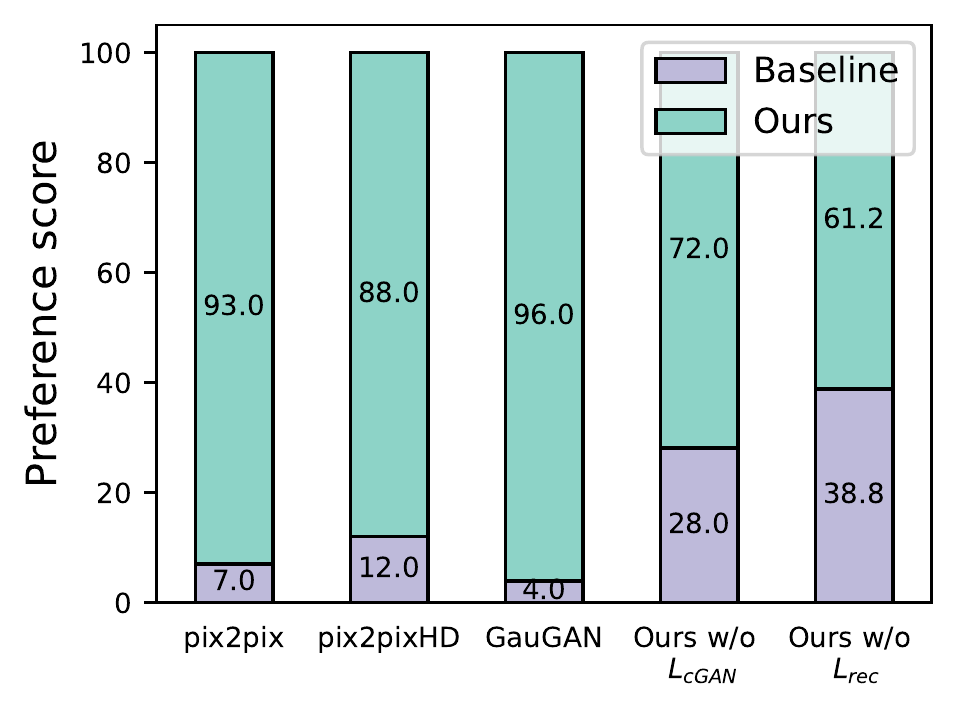}
    \vspace{-15pt}
    \caption{Bar plot for visual user study.}
    \lblfig{user_study_results_visual}
  \end{subfigure}
  \hfill
  \begin{subfigure}[b]{0.48\linewidth}
    \includegraphics[width=\linewidth]{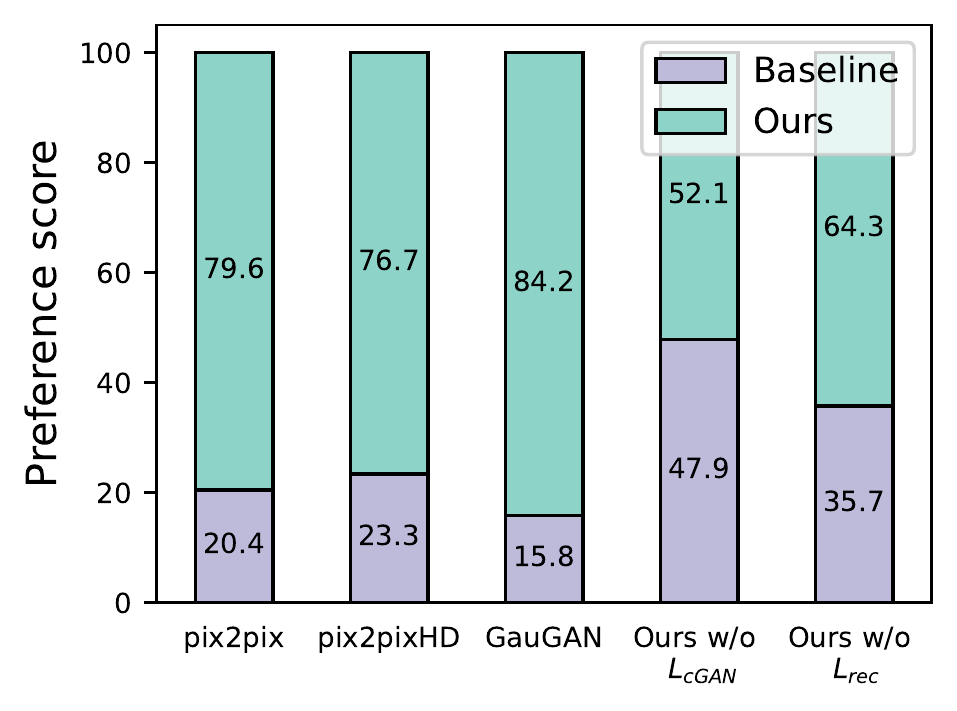}
    \vspace{-15pt}
    \caption{Bar plot for haptic user study.}
    \lblfig{user_study_results_haptic}
  \end{subfigure}
  \vspace{-5pt}
  \caption{\small {\bf Human perceptual study.} For each paired comparison, our method is preferred ($\geq$ 50\%) over the baseline for both visual and haptic output.
}
  \lblfig{user_study_results}
\end{figure}

%% file: figText/user_study_setup.tex
\begin{figure}[t]
\centering
\includegraphics[width=0.45\textwidth]{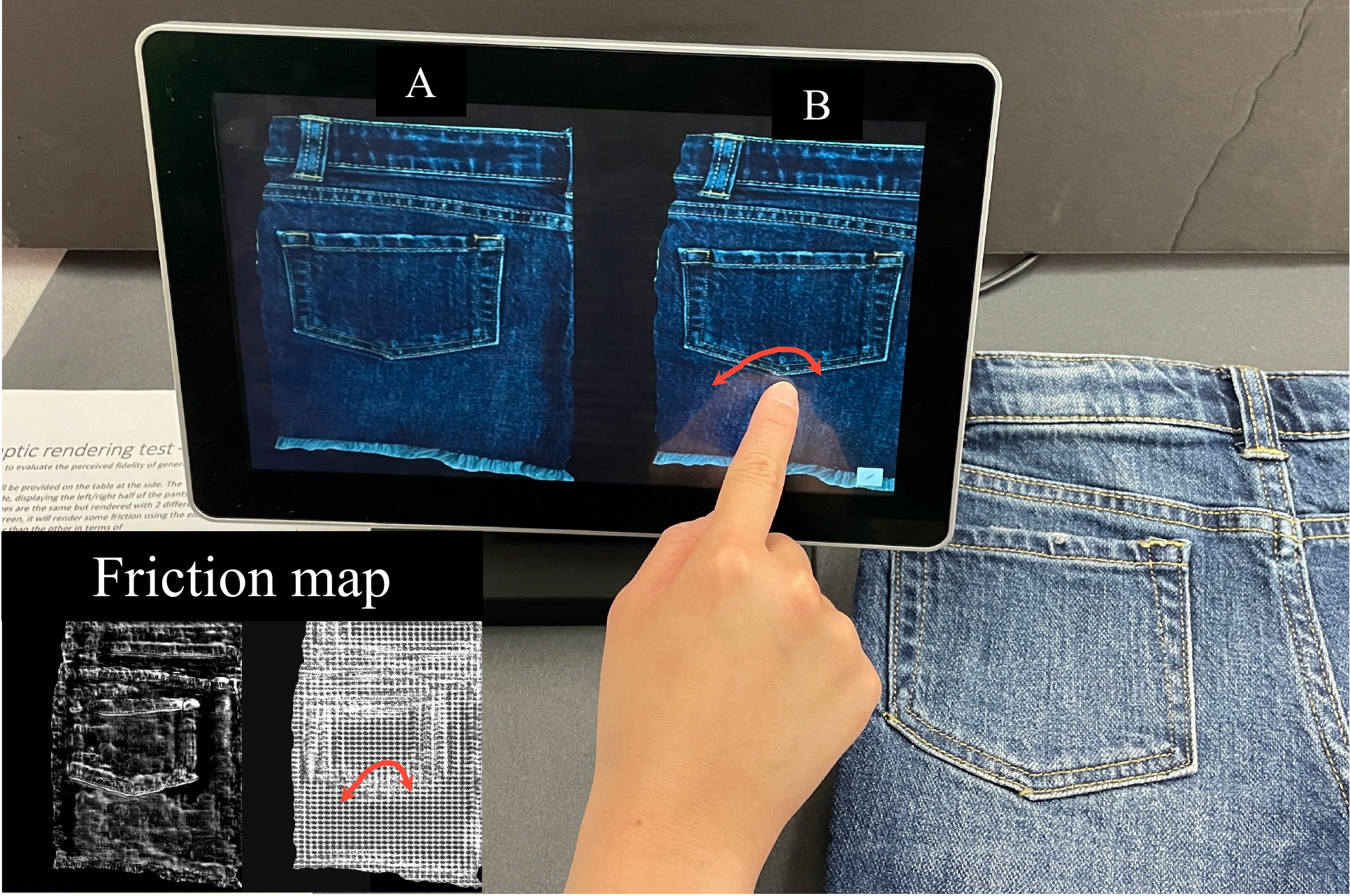}
\vspace{-5pt}
\caption{\small {\bf Experiment setup for the user study.} We perform an A/B test comparing the haptic output of our method and one of the baselines. The lower left corner shows the rendered haptic signal (friction map). The real garment is put on one side for reference. 
}
\lblfig{user_study_setup}
\vspace{-15pt}
\end{figure}

%% file: sections/6_conclusion.tex
\section{Discussion and Limitations}
\lblsec{discussion}

In this work, we presented a new method for automatically synthesizing visual and tactile images according to user inputs such as sketch and text. 
We used a high-resolution tactile sensor GelSight to capture the high-fidelity local geometry of objects. 
We then proposed a new conditional GAN model to generate visual and tactile output given a single sketch image. 
Finally, we introduced a pipeline to render visual and tactile outputs on the TanvasTouch touchscreen.

Our visual-tactile synthesis method can be used for different materials and objects, providing users with a more immersive experience when exploring virtual objects. 

\input{figText/failure_case}
\myparagraph{Limitations.} 
First, as shown in \reffig{failure_case}, distinctive patterns, such as enclosed letters, remain challenging. Our model fails to generalize to other user sketches. 
Second, as touch is an active perception, rendering performance relies on specific hardware constraints. 
In this work, the surface haptic device excels at rendering clothing, which is primarily flat with fine textures. 
Nevertheless, it is challenging to render 3D objects with substantial surface normal changes, such as an apple, on the same device.

Finally, since tactile data are collected during static touch and the rendering device mainly focuses on friction force, we can render roughness well but have limited capacity to render softness. 

\myparagraph{Societal impacts.}
Controllable visual-tactile synthesis for haptic rendering is a new research problem that has yet to be explored extensively.
We take the first step to address the modeling challenge and deploy our model to the latest hardware. 
Ultimately, we hope our work will facilitate multi-modal synthesis with generative models in applications such as online shopping, virtual reality, telepresence, and teleoperation.

%% file: figText/failure_case.tex
\begin{figure}[t]
\centering
\includegraphics[width=0.5\textwidth]{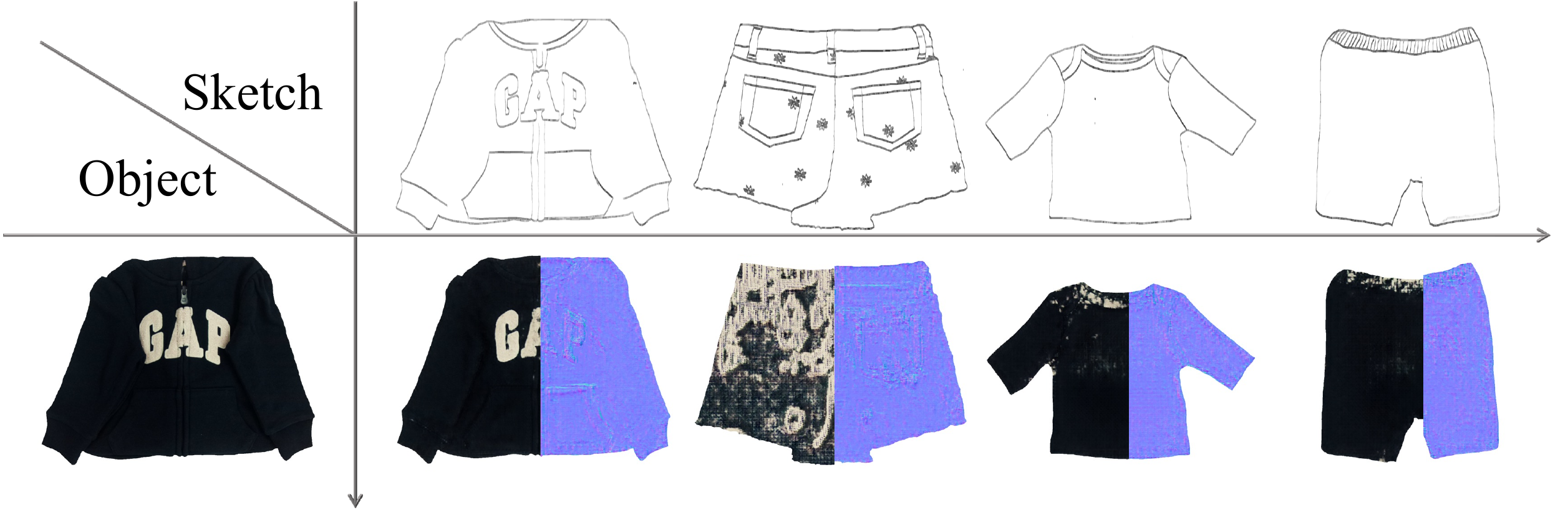}
\caption{\small {\bf Failure cases.} Our method struggles when the training object and testing sketch exhibit vastly different patterns. For example, given a hoodie with large letter patterns, our model reconstructs the training object (1st column) but fails to fill in the color within the flower pattern (2nd column) or to generate uniform texture around the neckline (3rd column) or the waistband (4th column).
}

\lblfig{failure_case}
\end{figure}

%% file: sections/7_ack.tex
\myparagraph{Acknowledgment.}
We thank Sheng-Yu Wang, Kangle Deng, Muyang Li, Aniruddha Mahapatra, and Daohan Lu for proofreading the draft. We are also grateful to Sheng-Yu Wang, Nupur Kumari, Gaurav Parmar, George Cazenavette, and Arpit Agrawal for their helpful comments and discussion. Additionally, we thank Yichen Li, Xiaofeng Guo, and Fujun Ruan for their help with the hardware setup. Ruihan Gao is supported by A*STAR National Science Scholarship (Ph.D.).

%% file: sections/8_supplement.tex
\noindent{\Large\bf Appendix}
\vspace{5pt}

In this appendix, we provide additional experimental results and details.

In \refsec{more_exp}, we present additional experimental details on our haptic perceptual study protocol, network architectures, and training procedures. 
In \refsec{more_results}, we show more results and analysis on our main method, additional ablation studies on each loss component, and comparison with a visual-tactile cross-modal prediction baseline. 
Please see our website video~\url{\mywebsitelink} for data capture and user interaction demos.

\section{Additional Experimental Details}
\lblsec{more_exp}

\myparagraph{Protocol for the haptic user study.}
The data are collected in person via psychophysical experiments, and please see our website video for a demo of user interaction and user study setup. 
Twenty people, 13 male and 7 female, with an average of 24.1 years (SD: 2.1), participated in the experiments. The experimental procedures have been approved by the Institutional Review Board (IRB) of the institution. All participants have provided informed consent and received compensation at 15 USD per hour.

Specifically, in each round, the participant is presented with a real garment at the side of the table and two rendering outputs on the TanvasTouch screen. 
The two renderings have the same visual appearance (ground-truth visual image) but different haptic outputs. 
One is generated by our method, and the other is generated by one of the baselines (pix2pix, pix2pixHD, and GauGAN). 

The participants are asked to slide their index finger of the dominant hand on the TanvasTouch screen and on the real clothes with any force and velocity as desired, freely switching back and forth. 
On the touchscreen, two rendering outputs are put side-by-side showing the same half of the clothes (left or right), and the participant is allowed to freely explore both sides and select one of them as more realistic within one minute.

Before each experiment, the participants are asked to complete a training session, which includes a brief introduction to the TanvasTouch device and a quick demo to render homogeneous textures provided by TanvasIntro App, an official demo designed by Tanvas Inc \textsuperscript{\textregistered}. 
These steps can help familiarize the participants with how the device works and what type of rendering feedback they would expect. 

We sample one garment object from the \datasetName dataset for a warm-up and leave the rest of the unseen objects for testing, following Richardson et al. ~\cite{richardson2022learning}. 
The warm-up and testing follow the same protocol described above, except that the warm-up session is not timed so the participants have enough time to explore the device. 
To prevent user fatigue, we randomly select 5 out of 19 unseen objects for each testing session and report the averaged results. 
Each experiment lasts approximately 60 minutes.

\myparagraph{Network architectures.} 
We use the notations in pix2pix~\cite{isola2017image} to describe our network architecture. 
Let \texttt{Ck} denote a Convolution-BatchNorm-LeakyReLU layer with $k$ filters, with the slope of 0.2 for LeakyReLU. 
\texttt{CTk} denotes a ConvolutionTranspose-BatchNorm-ReLU layer with a dropout rate of $50\%$. 
All convolutions are $4 \times 4$ spatial filters applied with stride 2. 
Convolutions in the encoder downsample by a factor of 2, whereas in the decoder, they upsample the feature maps by a factor of 2.

The encoder-decoder architecture consists of:\\
{\bf encoder:}\\
\texttt{C10-C20-C40-C80-C80-C80-C80-C80} \\
{\bf decoder:}\\
\texttt{CT80-CT160-CT160-CT160-CT160-CT80-CT40-CT20}

$\GI$ and $\GT$ share the encoder and the first four layers of the decoder.
After the last layer in the decoder, a convolution is applied to map the feature maps to the final output (3 channels for visual synthesis and 2 channels for tactile synthesis), followed by a Tanh function.
As an exception to the above notation, BatchNorm is not applied to the first \texttt{C10} layer in the encoder. 
We use U-Net architecture~\cite{ronneberger2015u}, with skip connections between each layer $i$ in the encoder and  layer $n-i$ in the decoder, where $n$ is the total number of layers.

The discriminator architecture follows pix2pix~\cite{isola2017image} and is \texttt{C64-C128-C256-C512}.
After the last layer, a convolution is applied to map the feature maps to a 1-dimensional output, followed by a Sigmoid function. 
As an exception to the above notation, BatchNorm is not applied to the first \texttt{C64} layer in the encoder. 

\myparagraph{Training details.}
We train a separate model for each garment in the dataset. 
To augment the dataset, we pad the visual and sketch images ($960\times 1280$) to $1800 \times 1800$ and randomly crop them into $1536 \times 1536$ while maintaining a closed contour for the sketch. 
We manually collect 300 tactile patches for each garment and collect 60,000 patches in total. We split them into 6:2:2 for train, validation, and test set.
We use Adam solver~\cite{kingma2014adam} with learning rate $lr=0.001$, $\beta_1=0$, and $\beta_2=0.99$ for optimization and use "encoding dim=4" for Spatial Positional Encoding stated in Sec 4.3 in the main text.
All experiments are conducted on a single 24GB NVIDIA RTX A5000 GPU. 
The training takes 16 hours, and the inference takes an average of 0.19s to simultaneously generate visual and tactile outputs of size $1536 \times 1536$.

\input{tblText/su_ablation_loss.tex}
\section{Additional Results}
\lblsec{more_results}

\myparagraph{More qualitative results for the main method.}
\Cref{fig:su_baseline_1,fig:su_baseline_2,fig:su_baseline_3,fig:su_baseline_4,fig:su_baseline_5} shows the qualitative results for our method compared with baselines for all 20 objects in the dataset. 
For each object, we show the ground truth image and input sketch on the leftmost column, followed by the visual and tactile output (shown as a 3D height map) generated by our method, pix2pix, pix2pixHD, and GauGAN, respectively.

\myparagraph{Additional ablation studies.}
Here we show qualitative results for our loss ablation studies in \reffig{ablLossAll}. 
As mentioned in Section 5 ``Ablation studies'' and Table 2 in the main paper, we ablate $\mathcal{L}_{\text{cGAN}}$ and $\mathcal{L}_{\text{rec}}$ on visual and tactile synthesis together. 
\reffig{ablLossAll} shows that qualitatively, removing adversarial loss produces overly smooth visual images and less coherent tactile output, while removing reconstruction loss leads to trivial tactile output and checkerboard artifacts in visual images. 
We show two examples, but a similar effect is observed for all objects in the dataset.

In addition, we also study the effects of adversarial and reconstruction loss separately for visual and tactile synthesis. 

Following the notations in Section 4.2 in the main text, we abbreviate the adversarial and reconstruction loss for visual and tactile synthesis, respectively, as follows, and ablate each component to study their effects on the final output. 
\begin{equation}
    \begin{aligned}
    \VcGAN &= V(\GI, \DI, \xinput, \I) + V(\GI, \DA, \xinput, \I) \\
    \TcGAN &= V(\GT, \DT, \xinput, \I, \T) \\
    \Vrec  &= \mathcal{L}_{\text{rec}}(\GI, \xinput, \I) \\
    \Trec  &= \mathcal{L}_{\text{rec}}(\GT, \Sp, \Tp) .
    \end{aligned}
\end{equation}

We show quantitative results in \reftbl{su_ablation_loss} and qualitative results in \reffig{ablLossAll}. 

\myparagraph{More qualitative results for testing on unseen sketches.}
Similar to Figure 7 in the main text, \reffig{su_swap} shows additional results of testing our trained models on unseen sketches. 
Our model can synthesize visual and tactile output for both seen and unseen sketches. 
\reffig{su_swap_visual} shows visual images and \reffig{su_swap_tactile} shows tactile output in the form of normal map.
Each column corresponds to an object material, and each row corresponds to a sketch input.
Outputs on the diagonal represent testing on the same object of training, while the off-diagonal items demonstrate that our model can generalize to unseen sketches.

\myparagraph{Cross-object model. } In our main paper, we need to obtain a separate model for each object. %
In contrast, we now train a cross-object model on a diverse set of objects and materials. %
To encode both visual appearance and tactile textures, we extract CLIP image embeddings~\cite{radford2021learning} from the ground truth visual image,  %
and use it to modulate the feature maps from the sketch input using Adaptive Instance Normalization (AdaIN)~\cite{huang2017arbitrary}. 
We fuse them at the latent bottleneck and each common layer of the decoder before it branches out.

\reffig{su_style_encoding_arch} shows the network architecture and \reffig{su_style_encoding_res} shows some qualitative results.
\input{figText/su_style_encoding_arch}
\input{figText/su_style_encoding_res}
Our cross-object model can synthesize reasonable visual and tactile outputs for seen material (in the 1st column). 
For unseen materials, it can synthesize coherent outputs for some objects, like the jeans (in the 2nd column), but fails to preserve the color for others.

\myparagraph{Other baselines.} 
Since there is little existing work on conditional visual-tactile synthesis, we also compare our method with VisGel~\cite{li2019connecting}, a recent work on visual-tactile cross-modal prediction. 
\input{figText/VisGel.tex}
VisGel proposes to synthesize plausible tactile signals from visual inputs and vice versa, inferring the visual image given the tactile signal using a conditional adversarial network. 
It trains a separate network for each direction (i.e., vision2touch and touch2vision). 
Their problem setting is \emph{different} from ours as we aim to synthesize visual and tactile outputs from a user sketch. 
The VisGel also takes videos as input while we take a single sketch. 

Nevertheless, we adopt the vision2touch network and train it with paired visual-tactile patches ($32 \times 32$). In VisGel's original setting,  the authors capture videos of a robot interacting with objects on a tabletop. %
It uses the first video frame and the first tactile reading (no interaction, no tactile contact) as the reference frame. To accommodate the difference in setup, we experiment with two variations. Both use the full visual image as the visual reference and the tactile image with no contact as the tactile reference. 

For the first variation \emph{VisGel (crop)}, we crop the visual patch from the full image and feed the patch as source input. 
For the second variation \emph{VisGel (soft-mask)}, we use the full visual image as source input, multiply it with a soft attention map, which has the highest value of 1 at the patch location, dilate it with a Gaussian blur, and use a minimum value of 0.1 for other locations. %
\reffig{VisGel} shows an example of the reference frame and source input frame for both variations and the original VisGel setting.

Unfortunately, the network training fails to converge to meaningful results for both variations. 
There are several potential reasons. 
First, the original work can learn an attention map to find contact points in visual images by tracking the robot's motion. 
Their input video sequence provides additional environment information to build up the correlation between visual and tactile patches. 
Our data only have a single object in the visual frame and thus making it hard for the network to understand the correspondence.
Second, the network in the original work is trained on a large-scale dataset (165 objects and 2.5M frames) and does not fit in our single-image dataset (1 object and $\sim$250 data points). 
More importantly, it does not allow us to synthesize both visual and tactile outputs given a user sketch. %

\input{figText/su_baseline.tex}

\input{figText/su_abl_loss_all.tex}

\input{figText/su_swap.tex}

%% file: tblText/su_ablation_loss.tex
\begin{table}[t]
\centering
\scalebox{0.8}{
\begin{tabular}{lccccc}
\toprule
\multirow{2}{*}{\textbf{Method} }          & \multicolumn{2}{c}{Visual} & \multicolumn{2}{c}{Tactile} \\
\cmidrule(lr){2-3}\cmidrule(lr){4-5}
 &  \text{LPIPS}$\downarrow$  &  \text{SIFID}$\downarrow$ &  \text{LPIPS}$\downarrow$  & \text{SIFID}$\downarrow$ \\ \midrule
Ours                            & 0.070  & 0.029 & \textbf{\textbf{0.676}} & 0.104 \\
Ours w/o $\VcGAN$   & 0.116 & 0.336 & 0.686 & 0.107 \\ 
Ours w/o $\TcGAN$   & 0.070 & 0.040 & 0.677 & 0.104 \\ 
Ours w/o $\Vrec$   & 0.078 & 0.073 & \textbf{0.676}  & \textbf{0.103} \\ 
Ours w/o $\Trec$   & \textbf{0.064} & \textbf{0.017} & 1.021 & 0.255 \\ 
\bottomrule
\end{tabular}%
}
\caption{\textbf{Ablation study of loss components}.  We isolate the adversarial loss and reconstruction loss for visual and tactile synthesis and study their individual effect on the output performance. We observe a trade-off between visual and tactile reconstruction, i.e., removing tactile reconstruction loss leads to slightly better visual synthesis performance; nevertheless, our full model achieves the best performance for visual-tactile synthesis.
}
\lbltbl{su_ablation_loss}
\end{table}

%% file: figText/su_style_encoding_arch.tex
\begin{figure}[t]
\centering
\includegraphics[width=0.5\textwidth]{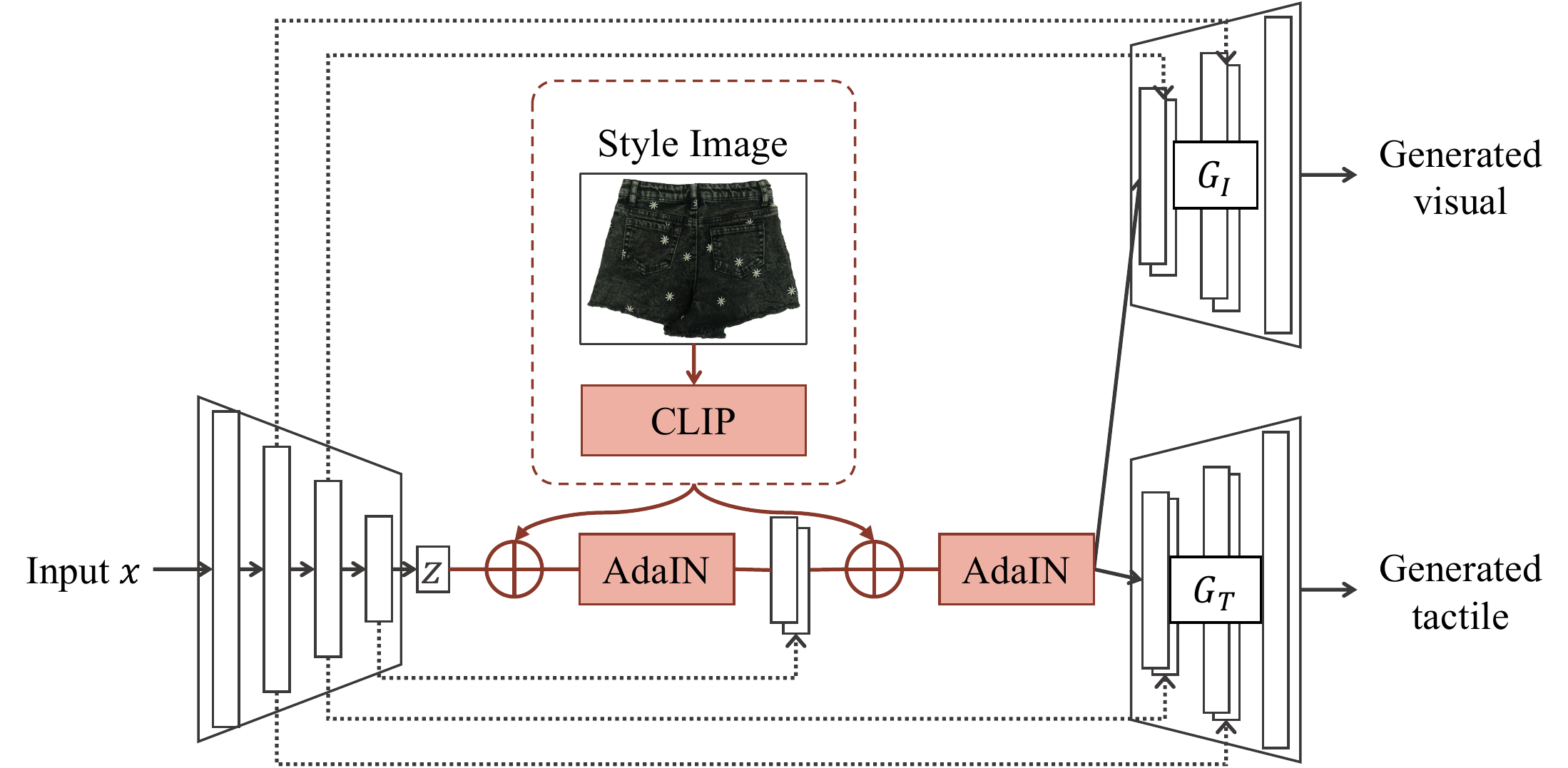}
    \caption{\small {\bf Cross-object model.} We extract a CLIP image encoding from an RGB image and integrate it with feature maps from sketch input using the AdaIN layer.
}

\lblfig{su_style_encoding_arch}
\end{figure}

%% file: figText/su_style_encoding_res.tex
\begin{figure}[t]
\centering
\includegraphics[width=0.475\textwidth]{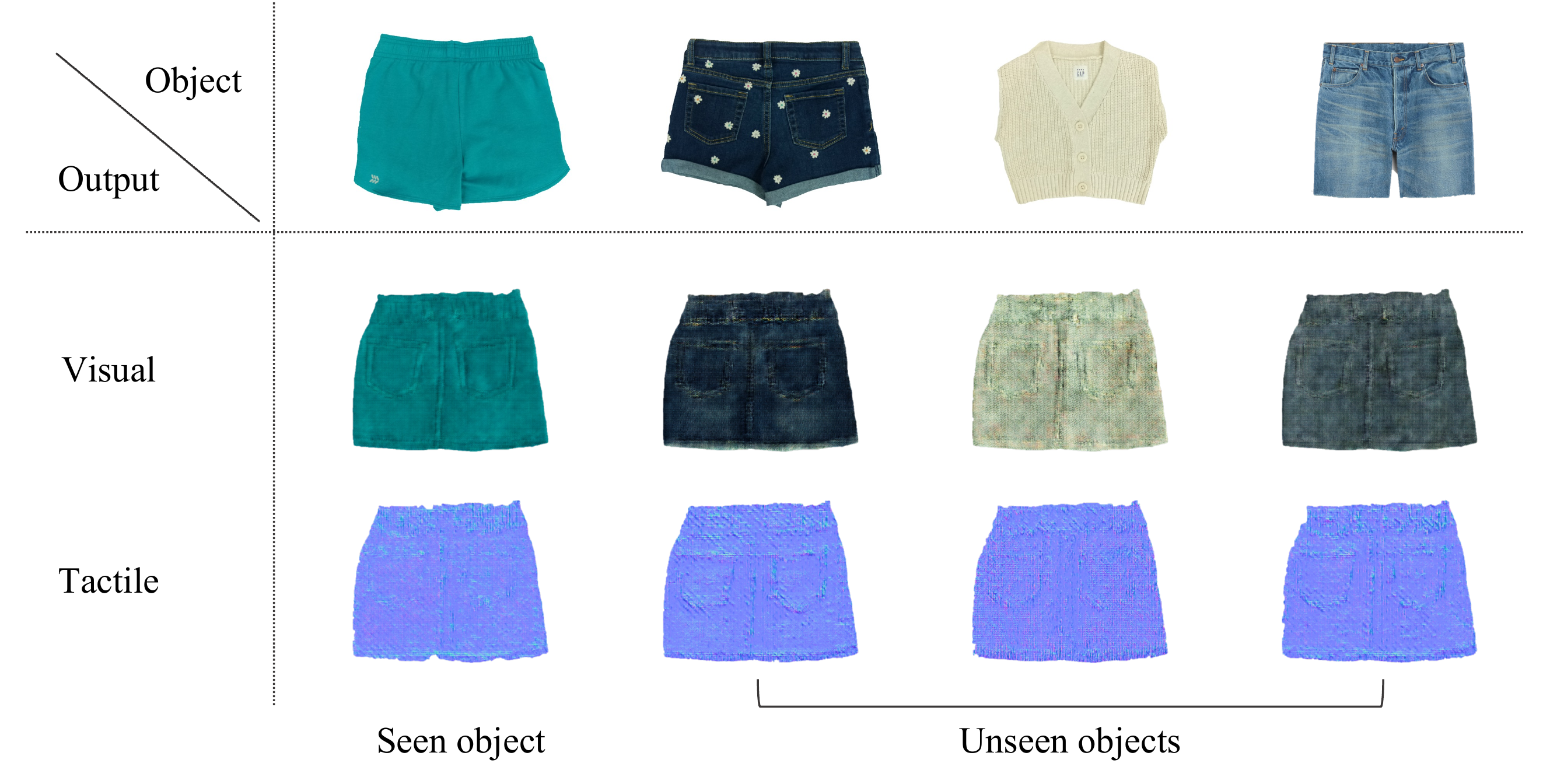}
    \caption{\small {\bf Cross-object model.} We train our model on 15 objects, and test with an unseen sketch with different style/material images. We show results on one seen image (1st column), two unseen images from our dataset (2nd and 3rd columns), and one online image  (4th column)
}

\lblfig{su_style_encoding_res}
\end{figure}

%% file: figText/VisGel.tex
\begin{figure}[t]
\centering
\includegraphics[width=0.50\textwidth]{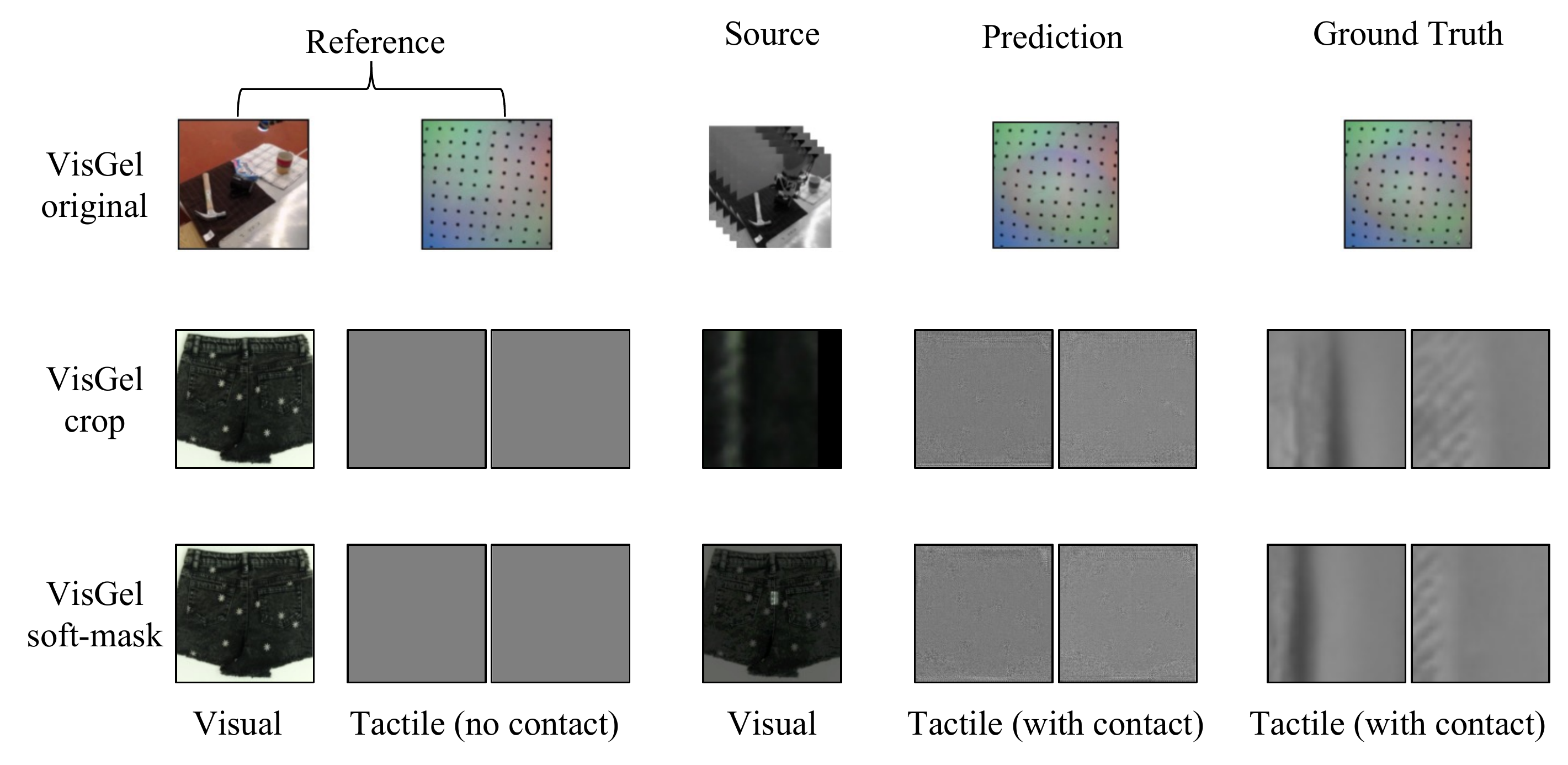}
    \caption{\small {\bf Variations of VisGel implementation.} We implement two variations, \textit{VisGel (crop)}, and \textit{VisGel (attention)}. The four columns show the reference frame (input), source image (input), prediction (output), and ground truth. Neither of the variations can provide comparative output as our method.
}

\lblfig{VisGel}
\end{figure}

%% file: figText/su_baseline.tex
\begin{figure*}[t]
  \centering
  \begin{subfigure}[b]{0.94\textwidth}
    \includegraphics[width=\linewidth]{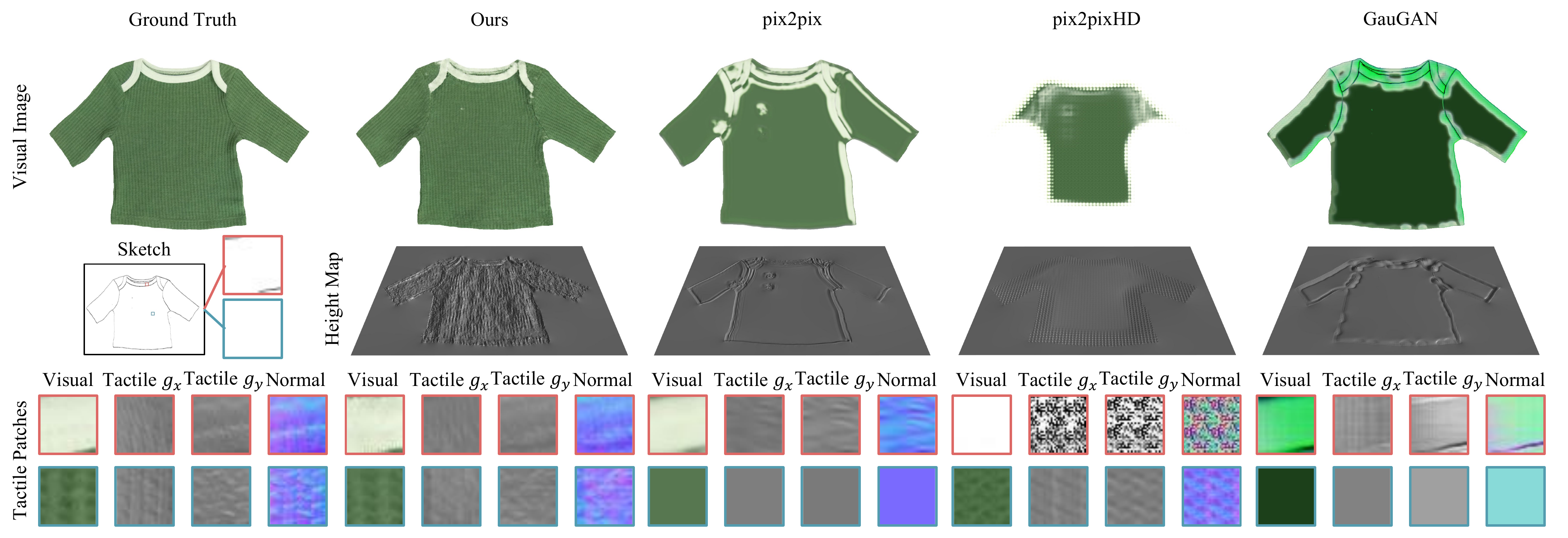}
    \vspace{-15pt}
  \end{subfigure}

  \begin{subfigure}[b]{0.94\textwidth}
    \includegraphics[width=\linewidth]{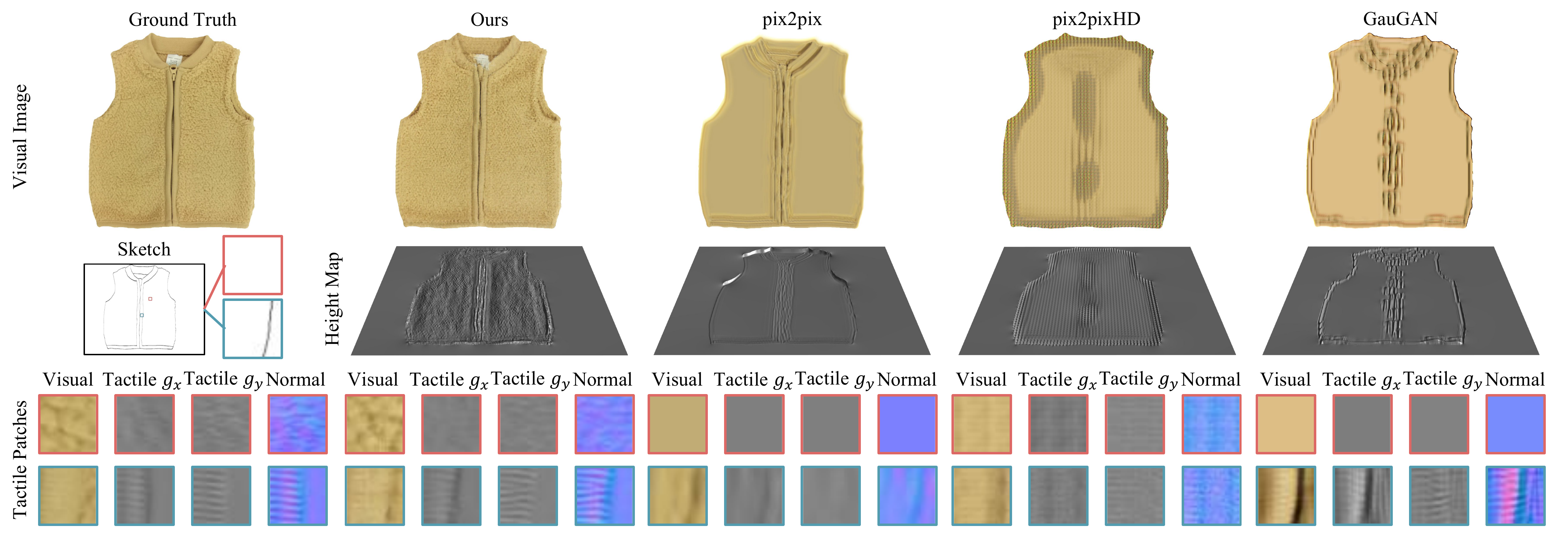}
    \vspace{-15pt}
  \end{subfigure}

  \begin{subfigure}[b]{0.94\textwidth}
    \includegraphics[width=\linewidth]{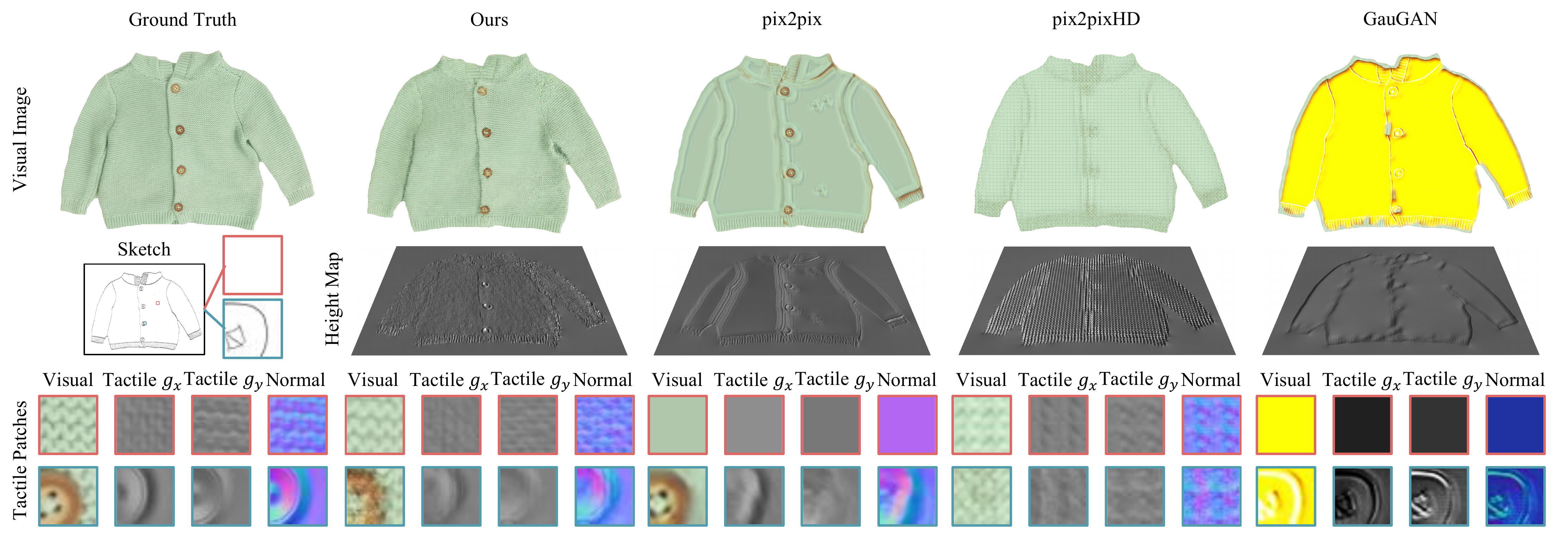}
  \end{subfigure}

  \begin{subfigure}[b]{0.94\textwidth}
    \includegraphics[width=\linewidth]{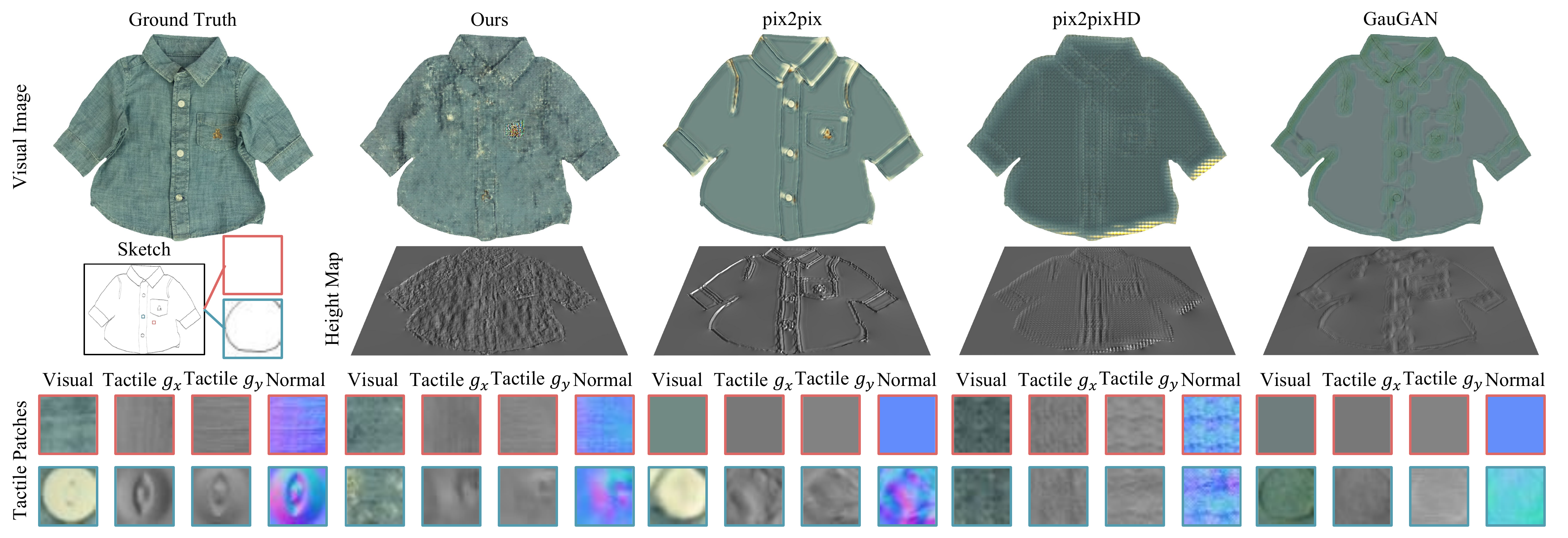}
  \end{subfigure}

  \caption{\small {\bf Qualitative results for our method compared with baselines for all 20 objects in the dataset.} For each object, we show the ground truth image and input sketch on the leftmost column, followed by the visual and tactile output (shown as a 3D height map) generated by our method, pix2pix, pix2pixHD, and GauGAN, respectively.
}
  \lblfig{su_baseline_1}
\end{figure*}

\begin{figure*}[t]
  \centering
  \begin{subfigure}[b]{0.94\textwidth}
    \includegraphics[width=\linewidth]{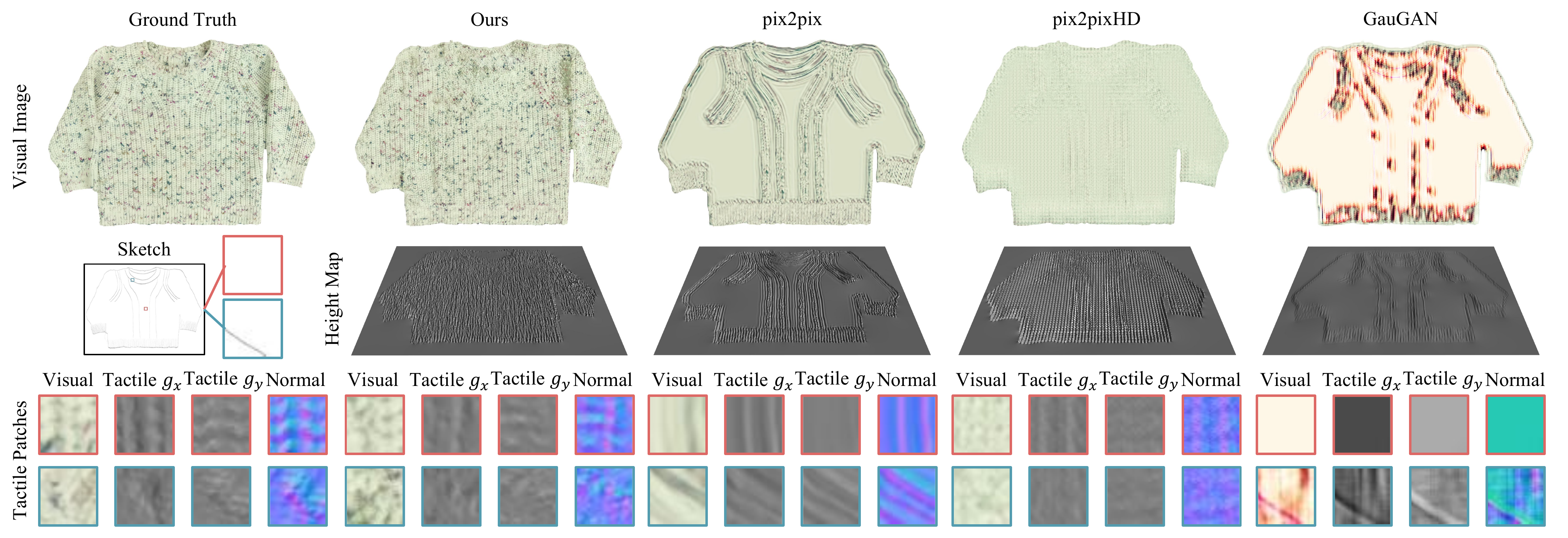}
    \vspace{-15pt}
  \end{subfigure}

  \begin{subfigure}[b]{0.94\textwidth}
    \includegraphics[width=\linewidth]{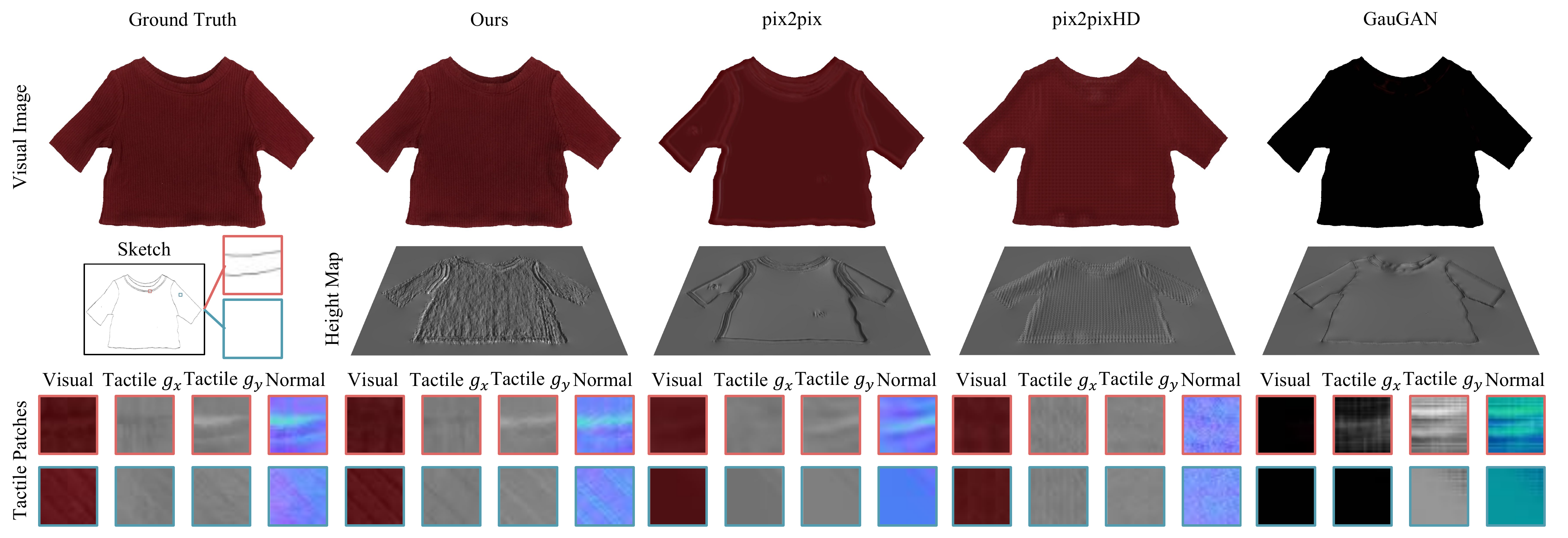}
    \vspace{-15pt}
  \end{subfigure}

  \begin{subfigure}[b]{0.94\textwidth}
    \includegraphics[width=\linewidth]{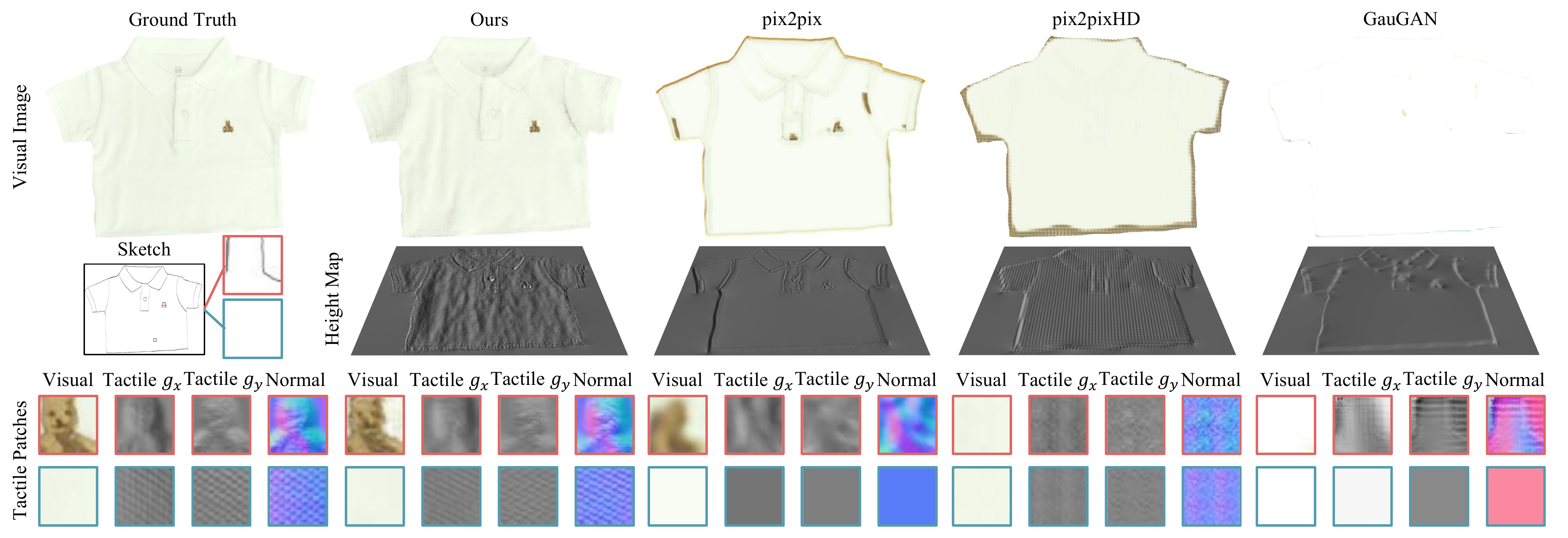}
    \vspace{-15pt}
  \end{subfigure}

  \begin{subfigure}[b]{0.94\textwidth}
    \includegraphics[width=\linewidth]{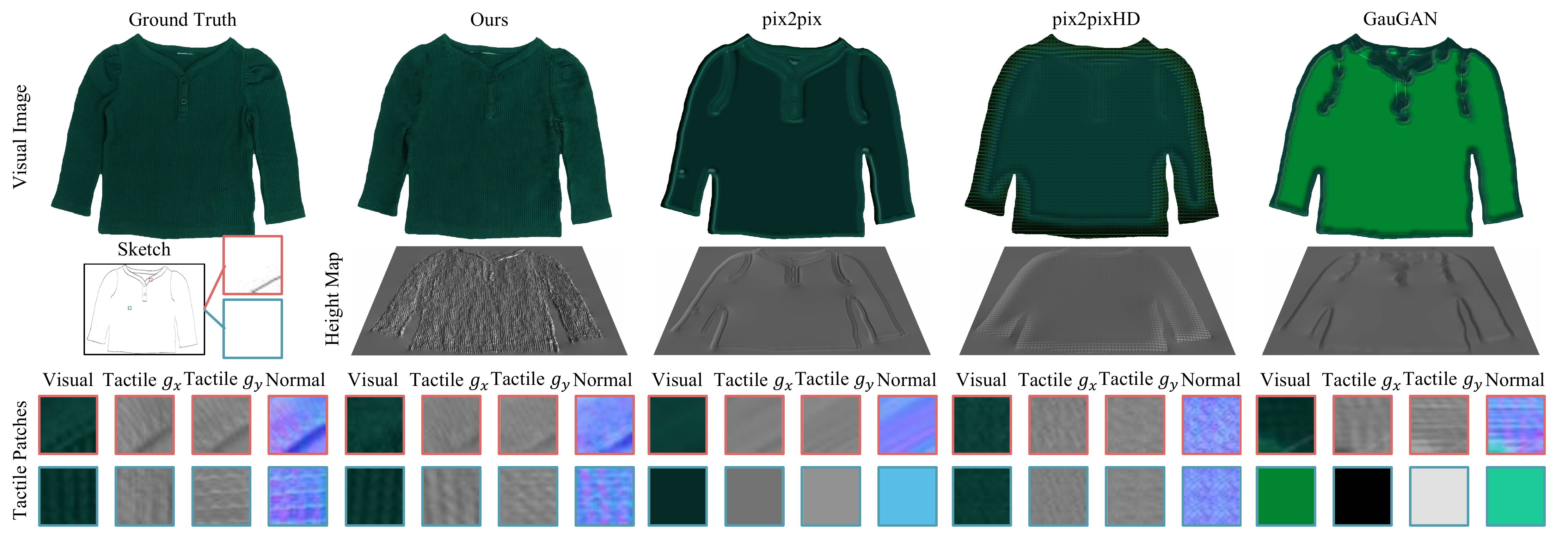}
    \vspace{-15pt}
  \end{subfigure}

  \vspace{-5pt}
  \caption{\small {\bf Qualitative results for our method compared with baselines for all 20 objects in the dataset (cont'd).}
}
  \lblfig{su_baseline_2}
  \vspace{-15pt}
\end{figure*}

\begin{figure*}[t]
  \centering
  \begin{subfigure}[b]{0.94\textwidth}
    \includegraphics[width=\linewidth]{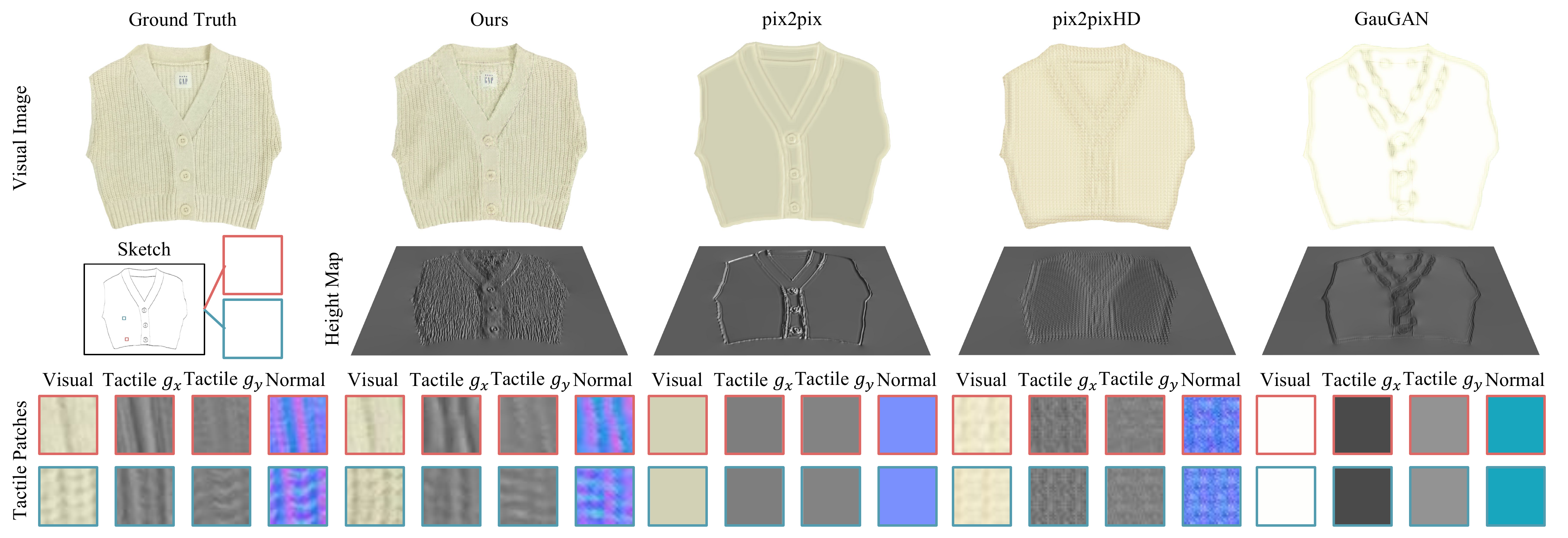}
    \vspace{-15pt}
  \end{subfigure}

  \begin{subfigure}[b]{0.94\textwidth}
    \includegraphics[width=\linewidth]{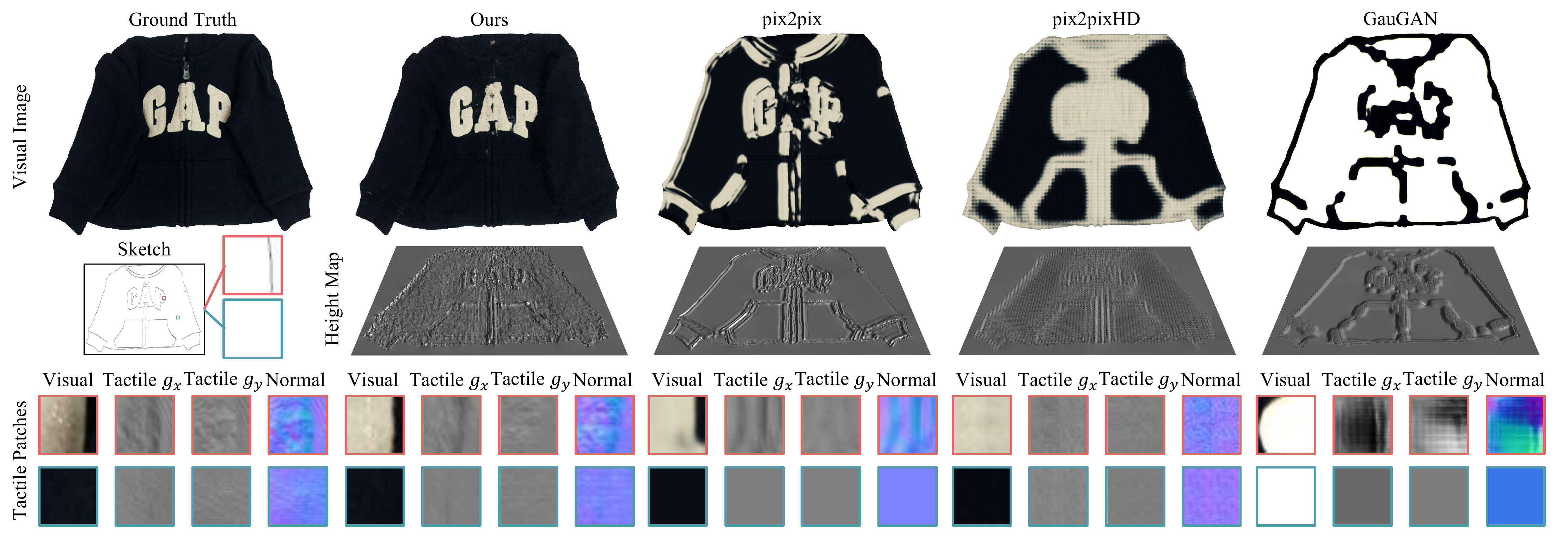}
    \vspace{-15pt}
  \end{subfigure}

  \begin{subfigure}[b]{0.94\textwidth}
    \includegraphics[width=\linewidth]{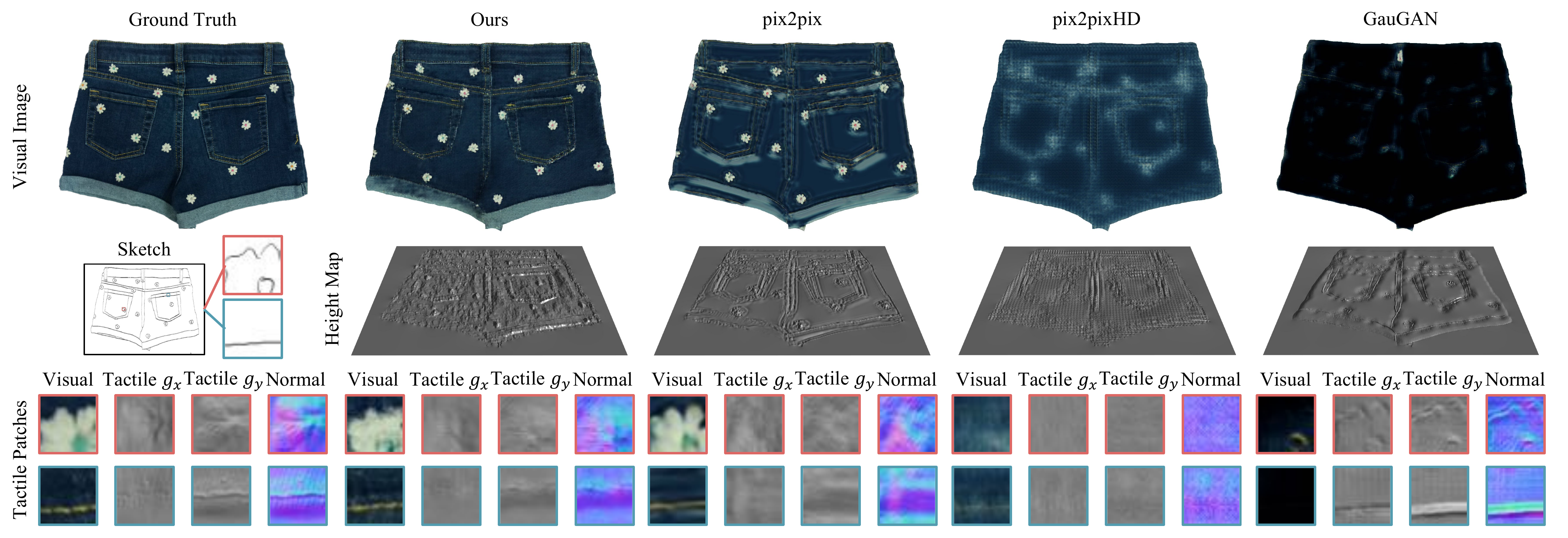}
    \vspace{-15pt}
  \end{subfigure}

  \begin{subfigure}[b]{0.94\textwidth}
    \includegraphics[width=\linewidth]{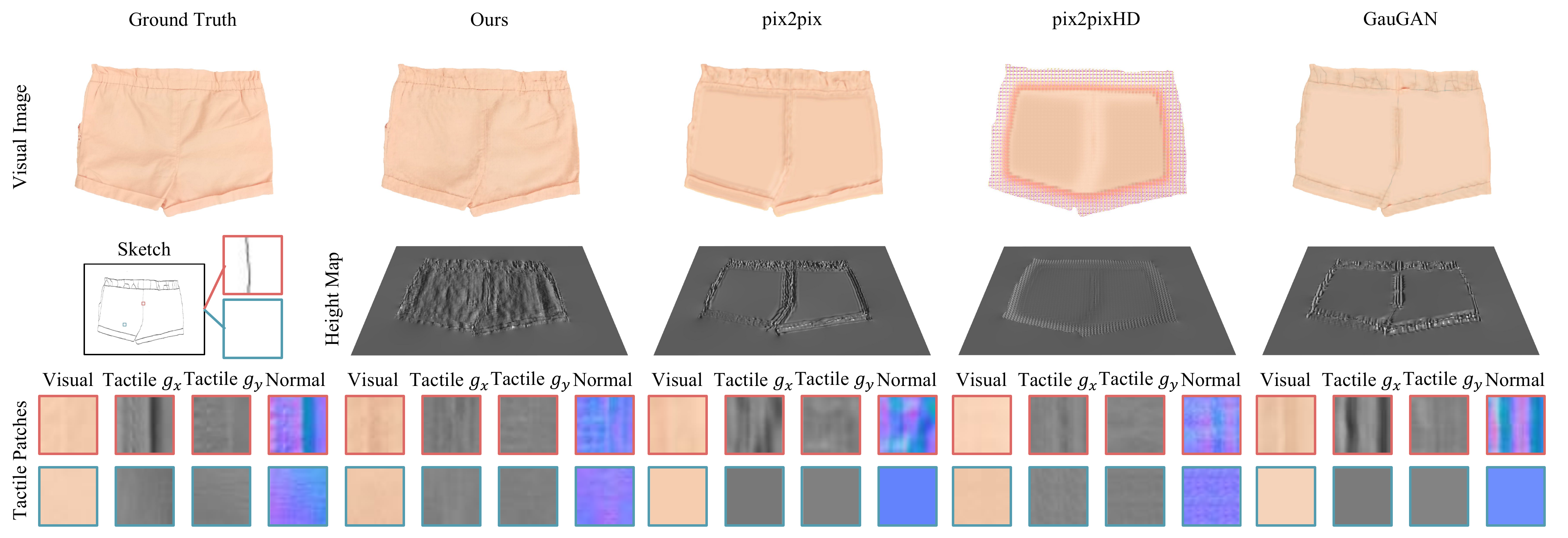}
    \vspace{-15pt}
  \end{subfigure}

  \vspace{-5pt}
  \caption{\small {\bf Qualitative results for our method compared with baselines for all 20 objects in the dataset (cont'd).}
}
  \lblfig{su_baseline_3}
  \vspace{-15pt}
\end{figure*}

\begin{figure*}[t]
  \centering
  \begin{subfigure}[b]{0.94\textwidth}
    \includegraphics[width=\linewidth]{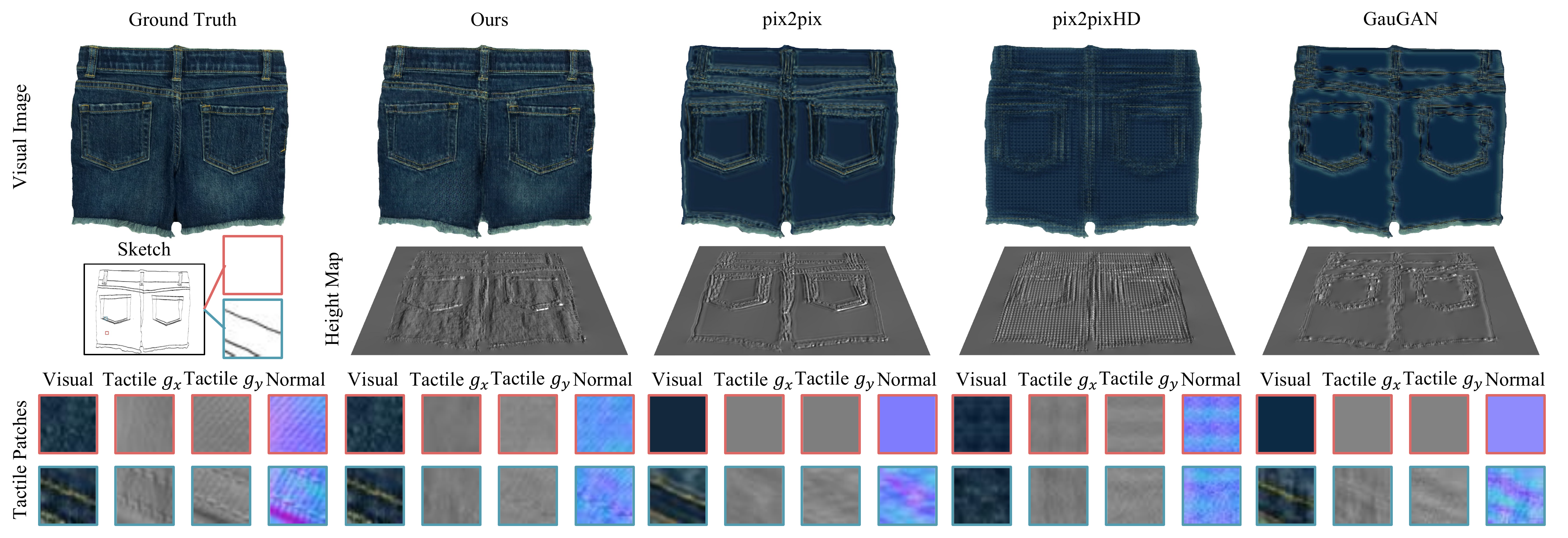}
    \vspace{-15pt}
  \end{subfigure}

  \begin{subfigure}[b]{0.94\textwidth}
    \includegraphics[width=\linewidth]{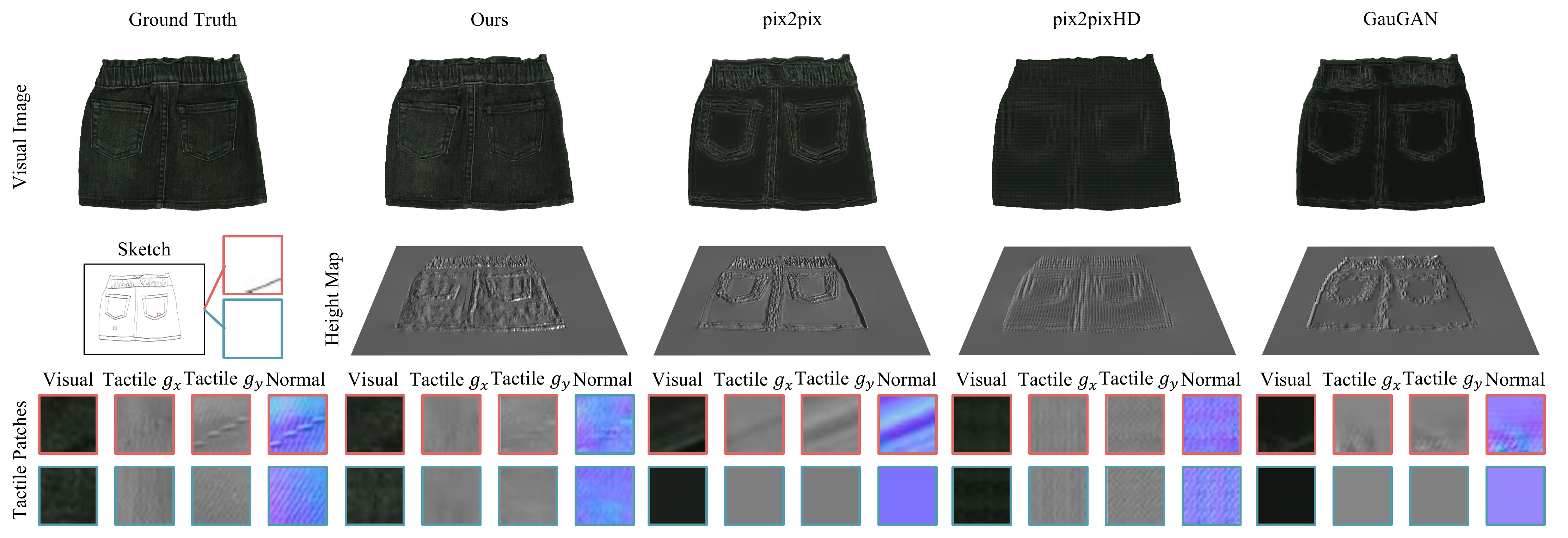}
    \vspace{-15pt}
  \end{subfigure}

  \begin{subfigure}[b]{0.94\textwidth}
    \includegraphics[width=\linewidth]{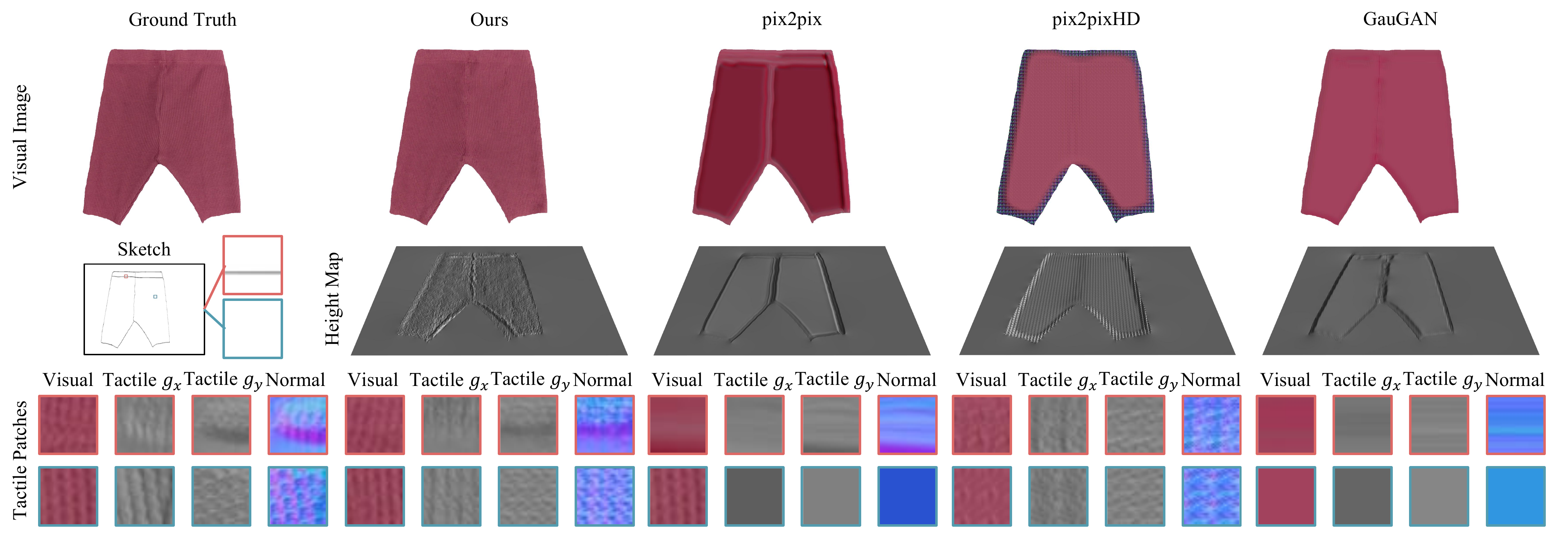}
    \vspace{-15pt}
  \end{subfigure}

  \begin{subfigure}[b]{0.94\textwidth}
    \includegraphics[width=\linewidth]{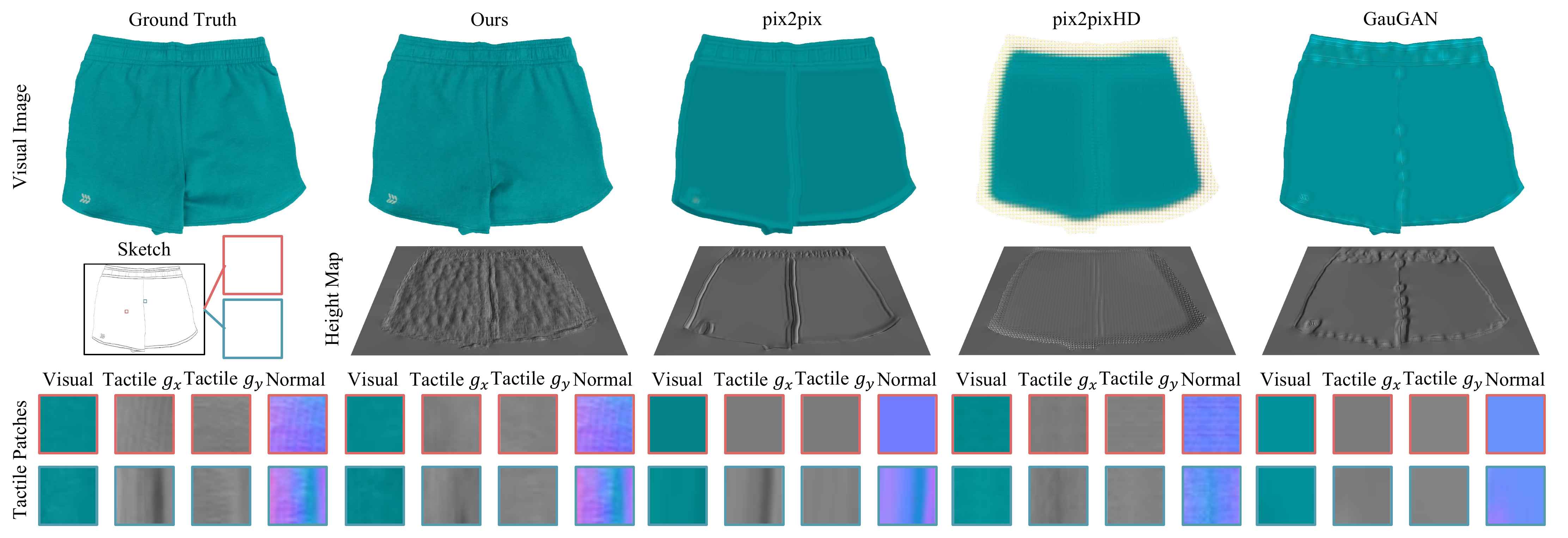}
    \vspace{-15pt}
  \end{subfigure}

  \vspace{-5pt}
  \caption{\small {\bf Qualitative results for our method compared with baselines for all 20 objects in the dataset (cont'd).}
}
  \lblfig{su_baseline_4}
  \vspace{-15pt}
\end{figure*}

\begin{figure*}[t]
  \centering
  \begin{subfigure}[b]{0.94\textwidth}
    \includegraphics[width=\linewidth]{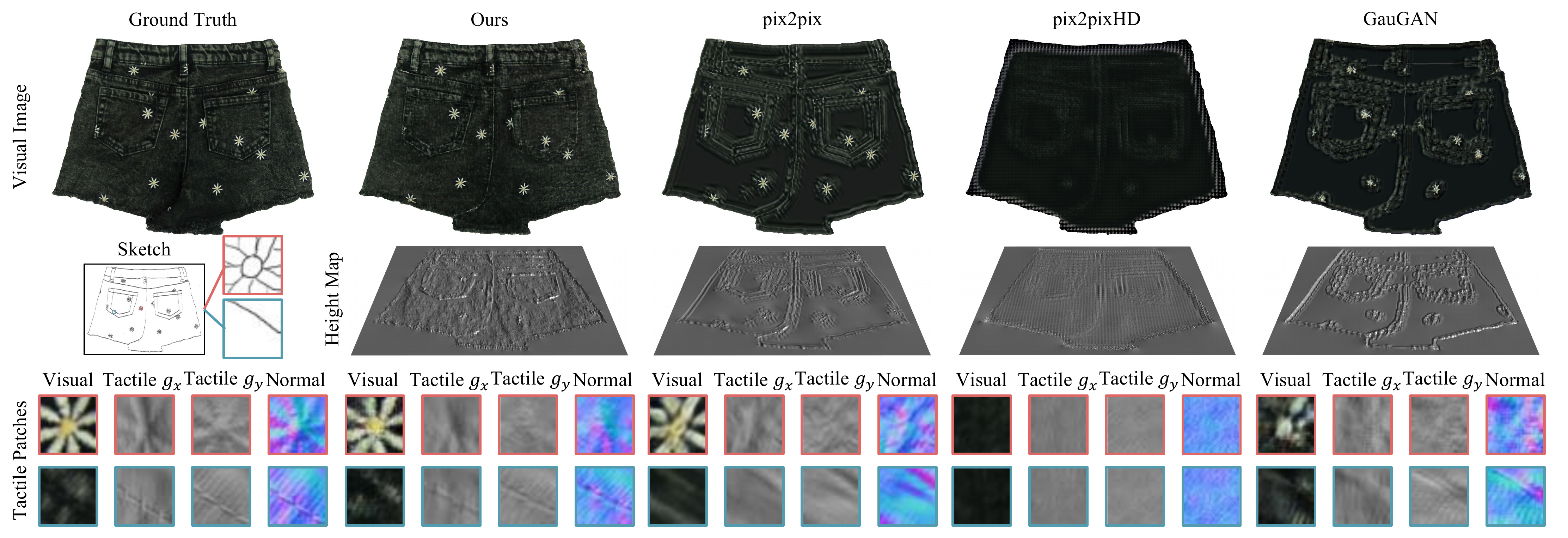}
    \vspace{-15pt}
  \end{subfigure}

  \begin{subfigure}[b]{0.94\textwidth}
    \includegraphics[width=\linewidth]{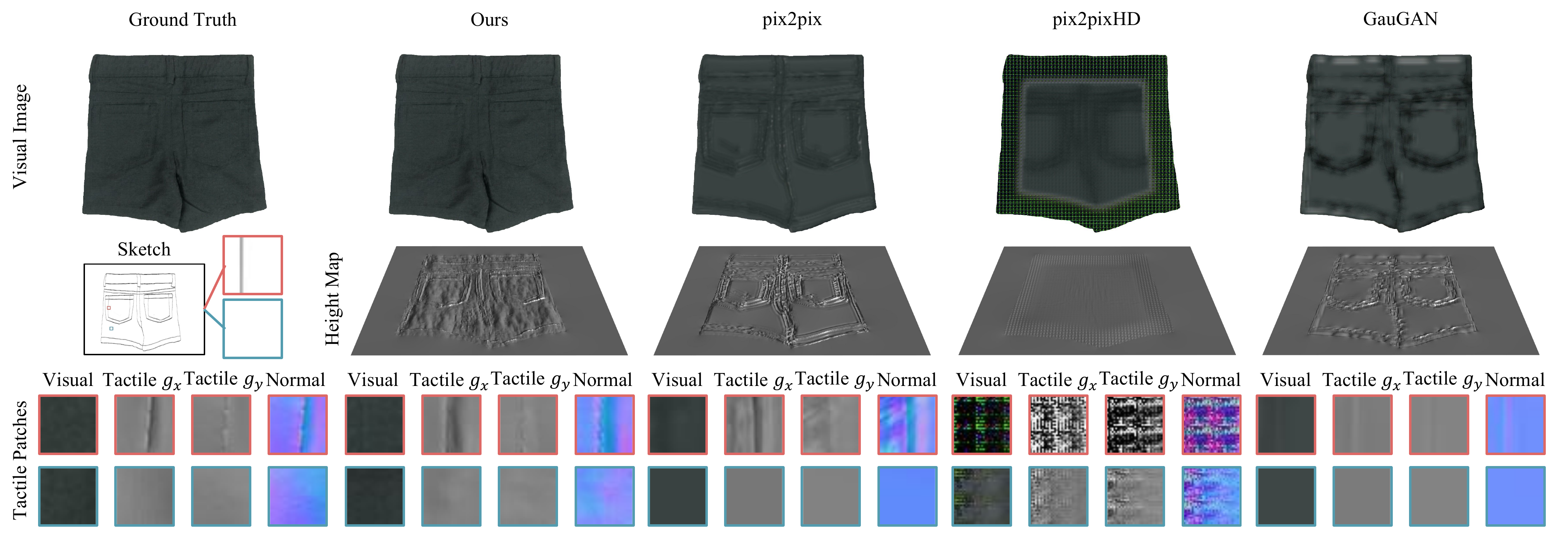}
    \vspace{-15pt}
  \end{subfigure}

  \begin{subfigure}[b]{0.94\textwidth}
    \includegraphics[width=\linewidth]{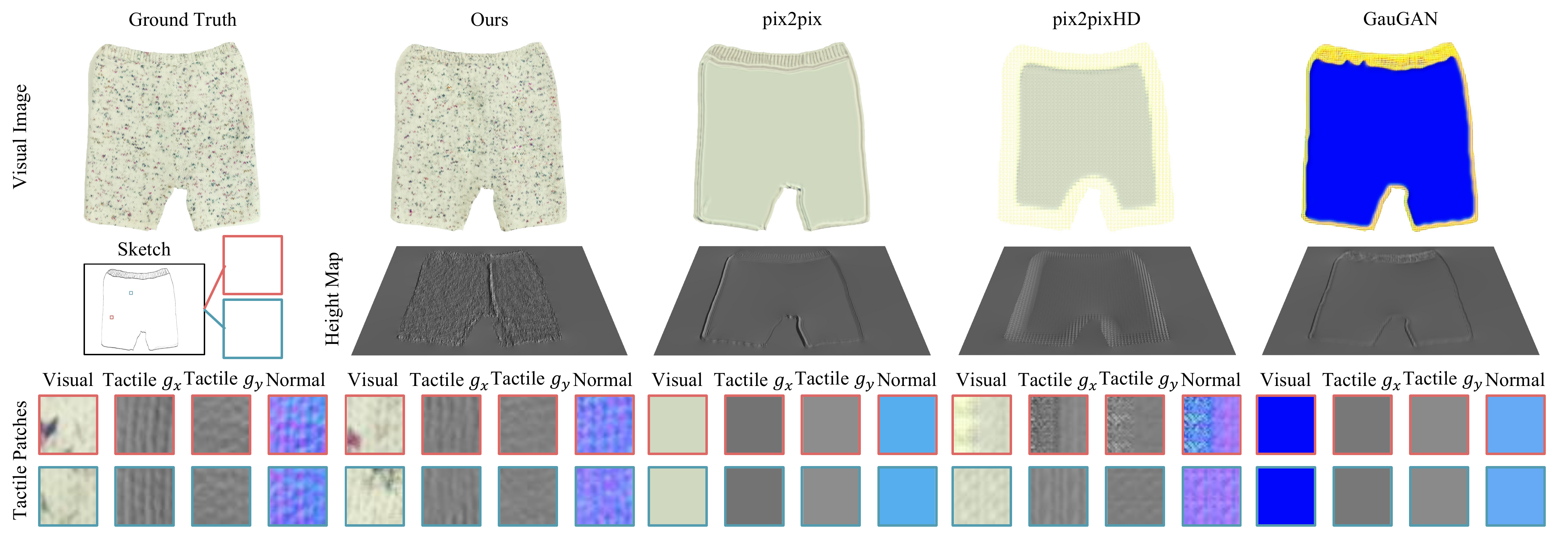}
    \vspace{-15pt}
  \end{subfigure}

  \begin{subfigure}[b]{0.94\textwidth}
    \includegraphics[width=\linewidth]{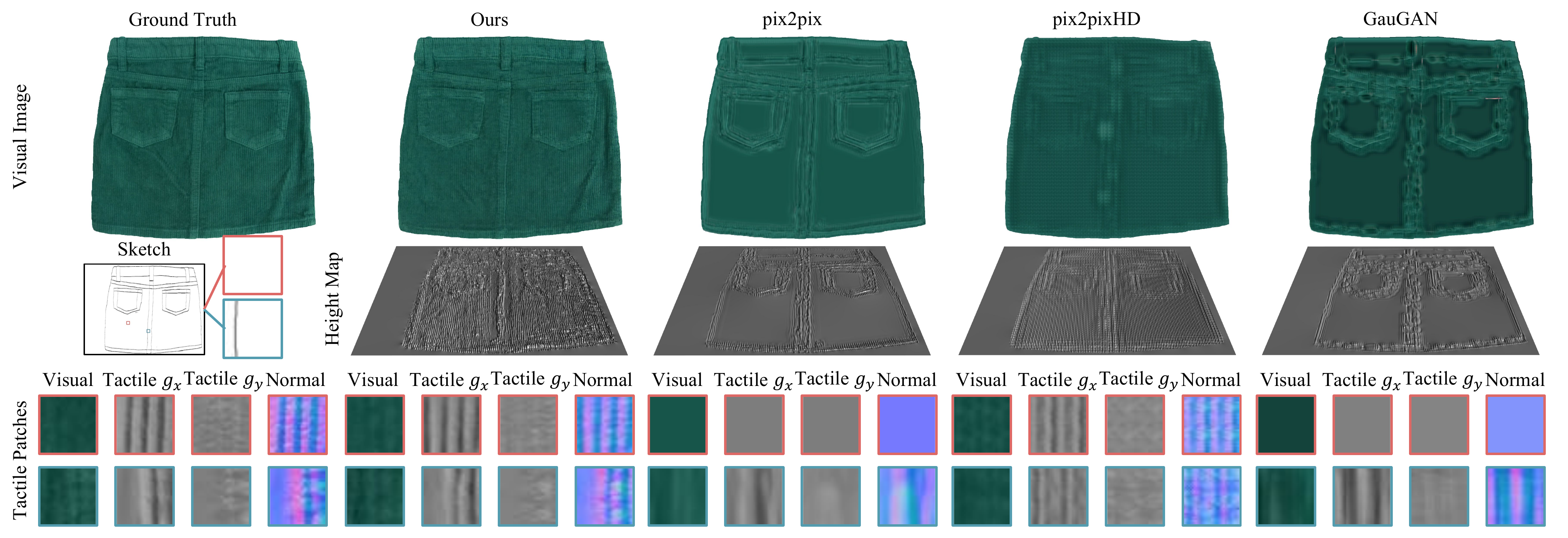}
    \vspace{-15pt}
  \end{subfigure}

  \vspace{-5pt}
  \caption{\small {\bf Qualitative results for our method compared with baselines for all 20 objects in the dataset (cont'd).}
}
  \lblfig{su_baseline_5}
  \vspace{-15pt}
\end{figure*}

%% file: figText/su_abl_loss_all.tex
\begin{figure*}[ht]
  \centering
  \begin{subfigure}[b]{0.67\textwidth}
    \includegraphics[width=\linewidth]{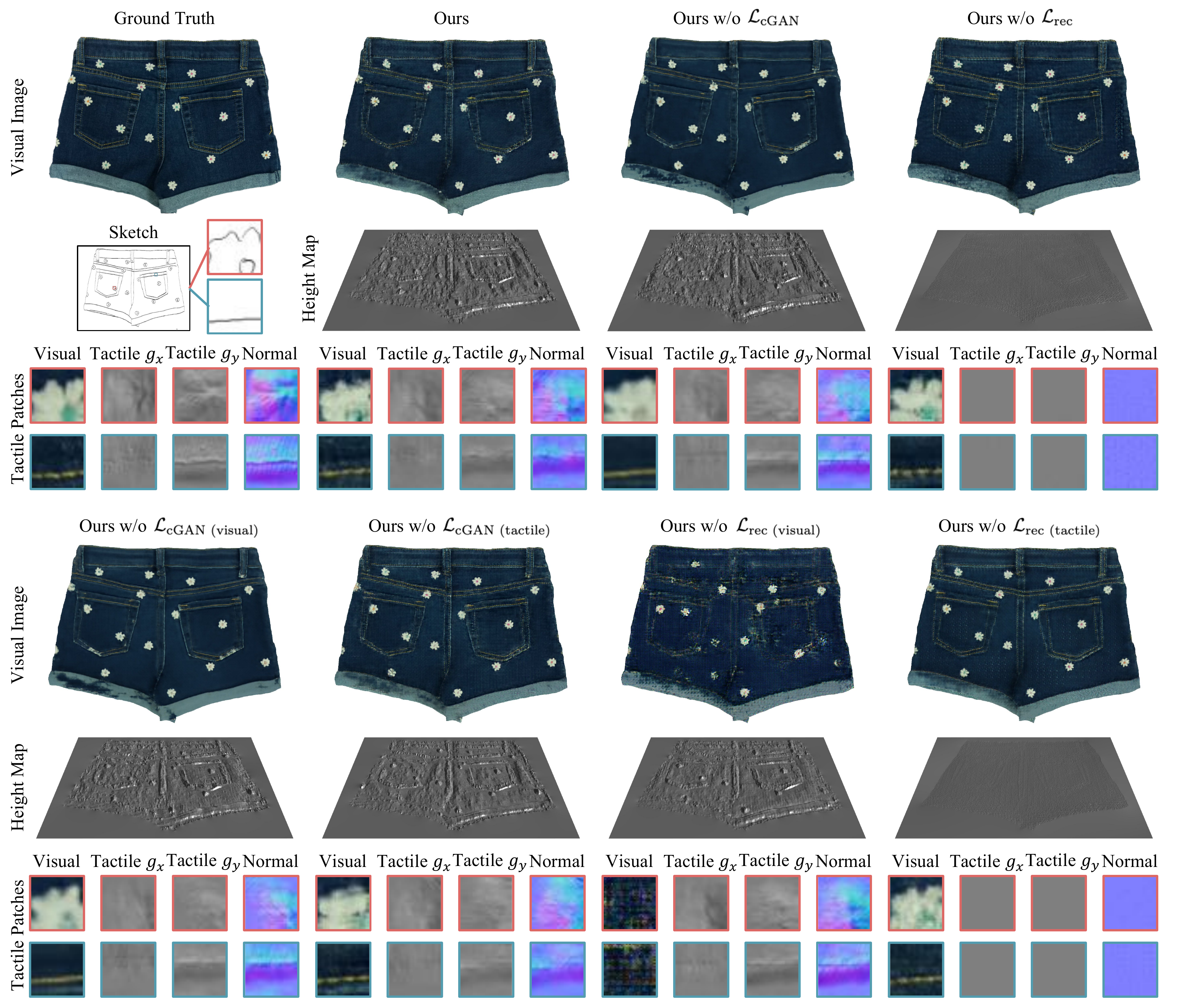}
    \caption{An example of a pair of jeans with flower-shaped decoration.}
  \end{subfigure}

  \vfill
  
   \begin{subfigure}[b]{0.67\textwidth}
    \includegraphics[width=\linewidth]{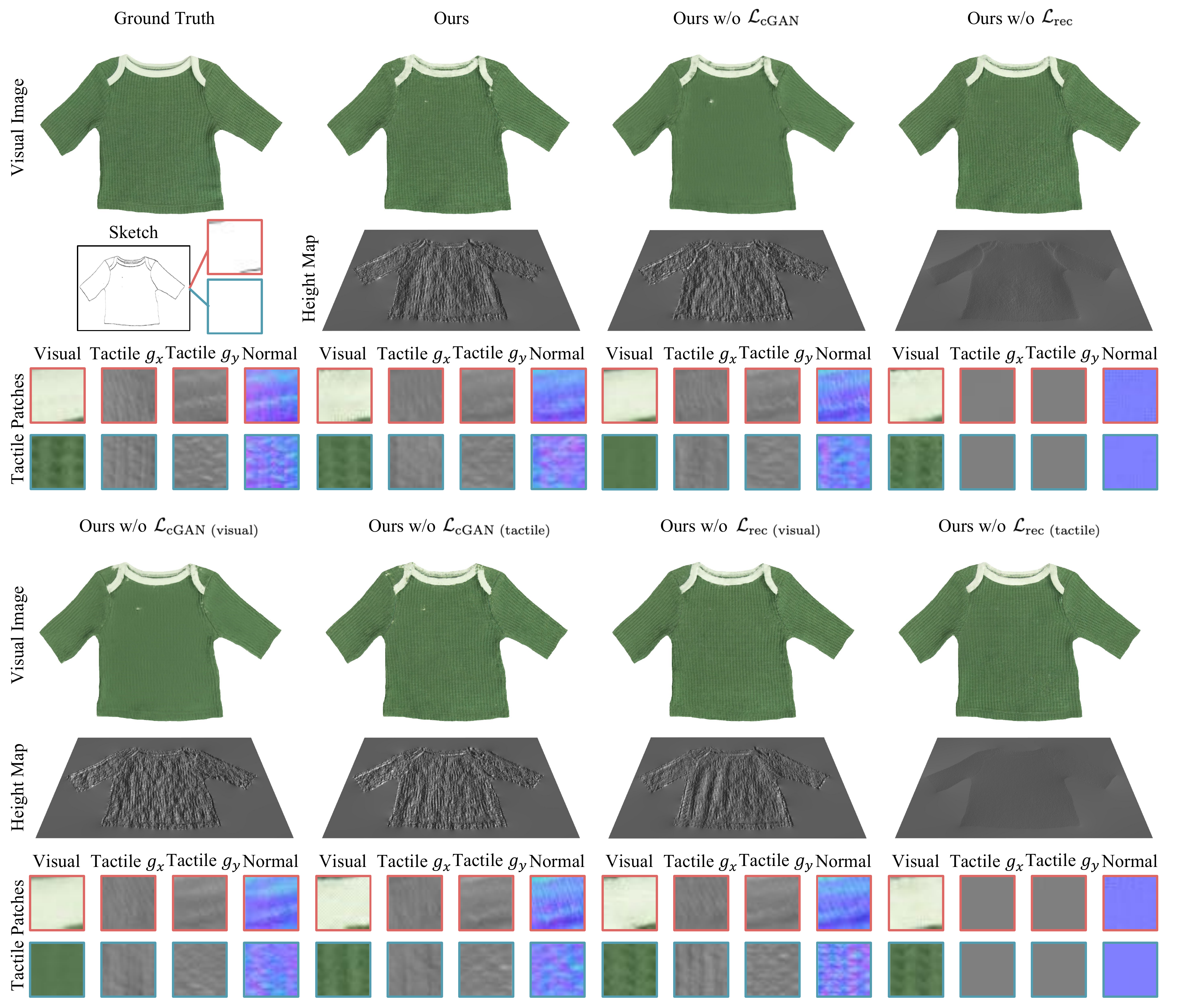}
    \caption{An example of a green shirt.}
  \end{subfigure} 

\caption{\small {\bf Qualitative comparisons for loss ablation studies.} We compare our method with six ablations, ours w/o $\mathcal{L}_{\text{cGAN}}$, ours w/o $\mathcal{L}_{\text{rec}}$, ours w/o $\VcGAN$, ours w/o $\TcGAN$, ours w/o $\Vrec$, and ours w/o $\Trec$.
For each method, we show, from top to bottom, the visual output, the tactile output (shown in heightmap), and two zoom-in patches. The input sketch to the network is shown in the upper leftmost column, with two color-coded bounding boxes indicating the location of two patches. In general, removing the adversarial loss results in overly smooth output, while removing reconstruction loss introduces trivial tactile output and visual artifacts.
}
\lblfig{ablLossAll}
\end{figure*}

%% file: figText/su_swap.tex
\begin{figure*}[ht]
  \centering
  \begin{subfigure}[b]{0.75\textwidth}
    \includegraphics[width=\linewidth]{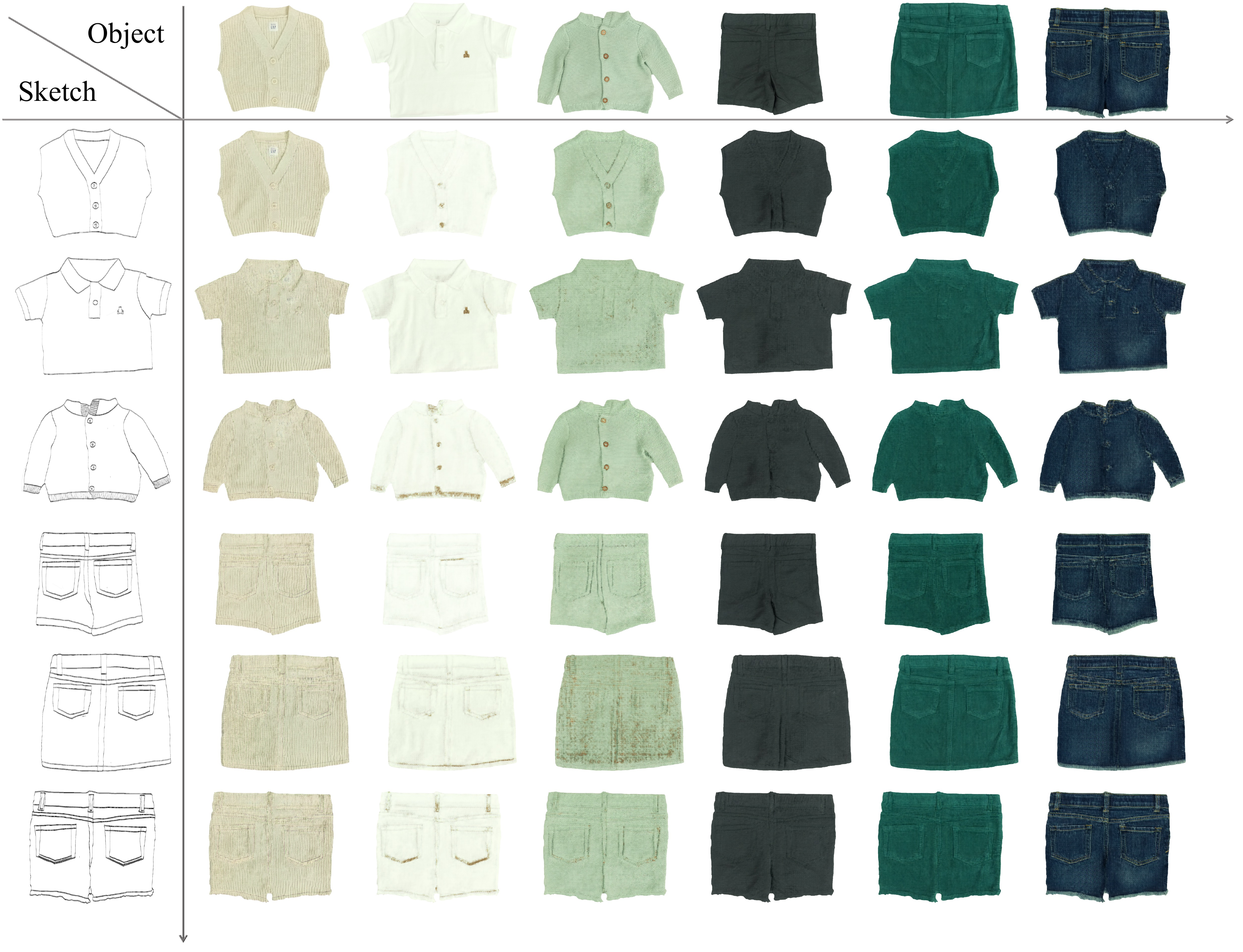}
    \caption{Visual output of testing on different sketches.}
    \lblfig{su_swap_visual}
  \end{subfigure}

    \vfill
    
   \begin{subfigure}[b]{0.75\textwidth}
    \includegraphics[width=\linewidth]{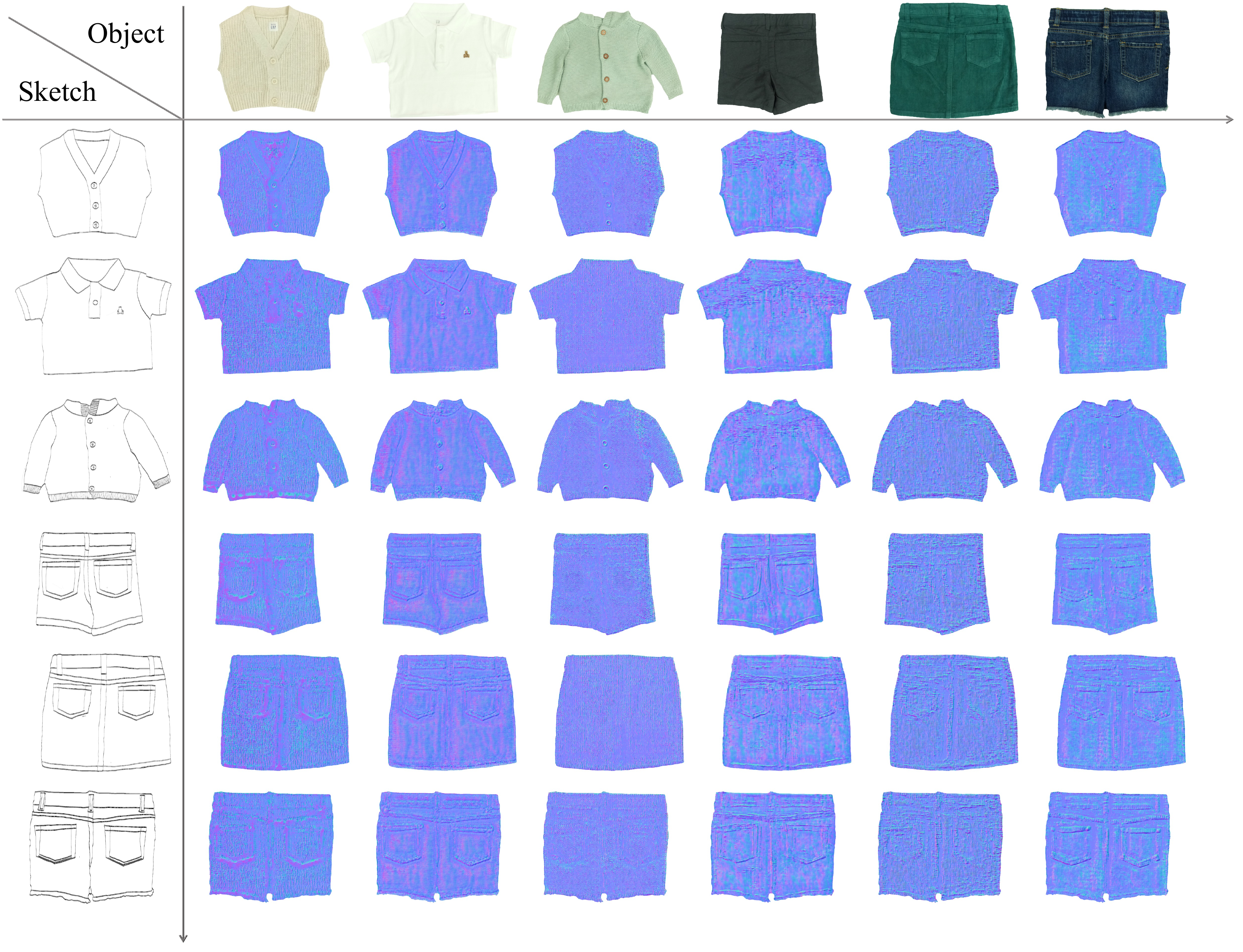}
    \caption{Tactile output of testing on different sketches (shown in format of normal map).}
    \lblfig{su_swap_tactile}
  \end{subfigure} 

\caption{\small {\bf Additional results on swapping sketch and material.} Our model can synthesize visual and tactile for both known and unseen sketches. For each sketch-material pair, we show the visual output in (a) and tactile output as a normal map in (b).
}
\lblfig{su_swap}
\vspace{-15pt}
\end{figure*}